\documentclass[lettersize,journal]{IEEEtran}
\usepackage[table]{xcolor}
\usepackage{amsmath,amsfonts}
\usepackage{algorithmic}
\usepackage{algorithm}
\usepackage{array}
\usepackage[caption=false,font=normalsize,labelfont=sf,textfont=sf]{subfig}
\usepackage{textcomp}
\usepackage{stfloats}
\usepackage{url}
\usepackage{verbatim}
\usepackage{graphicx}
\usepackage{cite}
\usepackage{times}
\usepackage{latexsym}
\usepackage{graphicx}
\usepackage{amsmath}
\usepackage{tabularx} % 用于创建可调整列宽的表格
\usepackage{booktabs} % 用于创建漂亮的表格线
\usepackage{multirow} % 用于合并单元格
\usepackage{leftidx}
\usepackage{hyperref}

\begin{document}

\title{Enhancing Human-like Multi-Modal Reasoning: A New Challenging Dataset and Comprehensive Framework}

\author{Jingxuan Wei, Cheng Tan, Zhangyang Gao, Linzhuang Sun,  Siyuan Li, \\
Bihui Yu, Ruifeng Guo, Stan Z. Li$^\dagger$,~\IEEEmembership{Fellow,~IEEE}
\thanks{Jingxuan Wei and Cheng Tan contributed equally as co-first authors. Gao, Sun, Li, Yu and Guo contributed to the project through selection, direction, code development, and writing assistance. Prof. Stan Z. Li is the corresponding authors.}
\thanks{Cheng Tan, Zhangyang Gao and Siyuan Li are with the AI Lab, Research Center for Industries of the Future, Westlake University, Hangzhou, China. Email: tancheng, gaozhangyang, lisiyuan@westlake.edu.cn}
\thanks{Jingxuan Wei,Linzhuang Sun are affiliated with the Shenyang Institute of Computing Technology, Chinese Academy of Sciences, Shenyang, China. Their institution of higher education is the University of Chinese Academy of Sciences, Beijing, China. Email: weijingxuan20, sunlinzhuang21@mails.ucas.edu.cn}
\thanks{Bihui Yu and Ruifeng Guo are with the Shenyang Institute of Computing Technology, Chinese Academy of Sciences, Shenyang, China. Email: yubihui@sict.ac.cn, guofuifeng@163.com.}
\thanks{Stan Z. Li is with the AI Lab, Research Center for Industries of the Future, Westlake University, Hangzhou, China. Email: stan.zq.li@westlake.edu.cn.}}

% \author{Cheng Tan, Zhangyang Gao,Stan Z. Li$^\dagger$ ,~\IEEEmembership{Staff,~IEEE,}

\markboth{Journal of \LaTeX\ Class Files,~Vol.~14, No.~8, August~2021}%
{Shell \MakeLowercase{\textit{et al.}}: A Sample Article Using IEEEtran.cls for IEEE Journals}


\maketitle

\begin{abstract}
Multimodal reasoning is a critical component in the pursuit of artificial intelligence systems that exhibit human-like intelligence, especially when tackling complex tasks. While the chain-of-thought (CoT) technique has gained considerable attention, the existing ScienceQA dataset, which focuses on multimodal scientific questions and explanations from elementary and high school textbooks, lacks a comprehensive evaluation of diverse approaches. To address this gap, we present COCO Multi-Modal Reasoning(COCO-MMR) dataset, a novel dataset that encompasses an extensive collection of open-ended questions, rationales, and answers derived from the large object dataset COCO. Unlike previous datasets that rely on multiple-choice questions, our dataset pioneers the use of open-ended questions in the context of multimodal CoT, introducing a more challenging problem that effectively assesses the reasoning capability of CoT models. Through comprehensive evaluations and detailed analyses, we provide valuable insights and propose innovative techniques, including multi-hop cross-modal attention and sentence-level contrastive learning, to enhance the image and text encoders. Extensive experiments demonstrate the efficacy of the proposed dataset and techniques, offering novel perspectives for advancing multimodal reasoning.  The data and code are available at \href{https://github.com/weijingxuan/COCO-MMR}{https://github.com/weijingxuan/COCO-MMR}.
\end{abstract}


\section{Introduction}
\IEEEPARstart{A}{dvancing} artificial intelligence systems towards human-like intelligence necessitates robust multimodal reasoning capabilities, especially when addressing complex tasks that require a holistic understanding of visual and textual information~\cite{DBLP:conf/nips/2022,tan2022hyperspherical,DBLP:journals/corr/abs-2201-08239,DBLP:journals/tcsv/ChengJWZ22,DBLP:journals/tcsv/HeP20}. In recent years, the chain-of-thought (CoT) technique~\cite{DBLP:conf/nips/LuMX0CZTCK22,DBLP:journals/corr/abs-2302-00923,DBLP:journals/corr/abs-2301-13379} has emerged as a promising approach for tackling multimodal reasoning challenges. CoT involves reasoning through a series of steps, where each step builds upon previous ones, mimicking the cognitive process humans employ when solving complex problems.

Despite the growing interest in CoT, the existing ScienceQA dataset~\cite{DBLP:conf/nips/LuMX0CZTCK22} comprising multimodal scientific questions and explanations extracted from elementary and high school textbooks has faced doubts due to the following reasons: (1) Limited scale: A relatively small scale of about 21k questions, which may limit the capability of learning models in capturing the diversity and complexity of reasoning. (2) Reliance on multiple-choice questions: Its reliance on multiple-choice questions may hinder the ability to generate open-ended responses that require coherent and comprehensive explanations. (3) Restriction to scientific question reasoning: The dataset is primarily focused on scientific question reasoning and lacks coverage of daily life scenarios. This confinement restricts the evaluation of models in broader real-world contexts.

To address these limitations and further advance the field of multimodal reasoning, we introduce COCO-MMR, a new dataset that is three times larger than ScienceQA, containing approximately 62k questions, rationales, and answers generated from the large object dataset COCO~\cite{lin2014microsoft}. One notable aspect of COCO-MMR is its departure from the convention of multiple-choice questions found in previous datasets. Instead, we introduce open-ended questions that require CoT models to generate detailed and coherent responses, offering a more challenging problem setting that better captures the reasoning capabilities of these models. Moreover, COCO-MMR is designed to cover diverse daily life scenarios, thereby extending the reasoning task into a broader real-world context beyond scientific questions. This helps evaluate whether models can apply commonsense knowledge and reasoning broadly to understand various concepts and situations in everyday life as humans do.

In addition to curating the comprehensive dataset, we propose a novel multimodal reasoning framework, Enigma-COT, that incorporates two innovative techniques: multi-hop cross-modal attention for boosting visual representation learning and sentence-level contrastive learning for strengthening the text encoder. The multi-hop cross-modal attention mechanism enhances the fusion of visual and textual information by considering multiple levels of interactions between different modalities. This enables models to capture more nuanced relationships and dependencies between visual and textual elements, leading to improved reasoning performance. Moreover, our sentence-level contrastive learning technique encourages the text encoder to capture finer-grained semantic relationships within sentences, enhancing the model's ability to reason based on subtle linguistic cues and textual context.

We conduct extensive experiments and employ a diverse set of metrics to ensure a fair and insightful comparison, thus effectively evaluating the effectiveness of our dataset and proposed techniques. The comprehensive experiments and detailed analysis we undertake provide compelling evidence for the efficacy of the COCO-MMR dataset and the proposed techniques. The results not only showcase the advantages of our approach over existing datasets but also offer novel insights into the realm of multimodal reasoning. By demonstrating the superior performance of our proposed method, we pave the way for further advancements in multimodal reasoning. Our work contributes to the development of models that exhibit enhanced reasoning capabilities across a wide range of real-world scenarios. We hope that these findings may propel the field forward, inspiring future research and innovation in multimodal reasoning.

\section{Related Work}

\subsection{Visual Question Answering}

Visual Question Answering (VQA) presents a challenging task that necessitates a model's comprehension of both natural language questions and their associated images, ultimately generating appropriate text responses~\cite{DBLP:journals/tcsv/XieFCHL22,DBLP:journals/tcsv/WangWLLZW23}. Situated at the intersection of computer vision~\cite{DBLP:journals/tcsv/DavaniS21,gao2022simvp,tan2022simvp,cao2022survey,DBLP:journals/tcsv/JinZFZLLZ22,zheng2023cvt} and natural language processing (NLP), VQA was initially proposed by~\cite{DBLP:conf/iccv/AntolALMBZP15}. A key challenge in VQA lies in the model's ability to understand both natural language queries and visual information, merging these distinct types of data to produce accurate answers.

Early VQA systems traditionally treated the question and image separately~\cite{WU201721,DBLP:journals/tmm/YuZLQHTW20}, integrating their representations in later stages to generate answers~\cite{DBLP:conf/iccv/AntolALMBZP15,DBLP:conf/icip/ChowdhuryNFS17,DBLP:conf/icip/HuangKJLJT18}. However, these approaches often overlooked the intrinsic interaction between the question and the image. To tackle this limitation, recent research has adopted advanced multimodal fusion strategies, including attention mechanisms, to more effectively integrate question and image information~\cite{DBLP:journals/tmm/Al-HalahG21,DBLP:journals/tcsv/OuCW21,DBLP:conf/miccai/SeenivasanIKR22,DBLP:journals/tcsv/TangLJPZDK23}. Nonetheless, these methods primarily focus on direct VQA tasks and may neglect challenges involving deep scientific knowledge and complex reasoning.

\subsection{Multimodal Reasoning}

Multimodal reasoning ~\cite{DBLP:journals/frai/WellsB21,DBLP:journals/tcsv/WenP21,DBLP:journals/jair/RasXGD22,DBLP:journals/tmm/WuLZLZQC23,DBLP:journals/tcsv/LiMDFT23} have garnered significant attention in reasoning tasks. These concepts revolve around understanding the decision-making process of models and generating interpretable reasoning paths. \textit{Lu et al.} ~\cite{DBLP:conf/nips/LuMX0CZTCK22} proposed the ScienceQA dataset and introduced a method based on reasoning chains. This work attempts to generate explainable reasoning chains with their scientific question dataset, and reveals a multi-step reasoning process to produce coherent reasoning chains when deriving the correct answer.  \textit{Zhang et al.} ~\cite{DBLP:journals/corr/abs-2302-00923} was the first to work on multimodal reasoning chains and utilize the multimodal information of the ScienceQA dataset by separately generating reasoning chains and answering in two stages. LLaMA-Adapter~\cite{gao2023llamaadapterv2} presents a lightweight adaptation method to efficiently finetune LLaMA~\cite{touvron2023llama} to achieve competitive multimodal reasoning performance with extremely low computational cost.

While previous approaches have demonstrated strong performance on the ScienceQA dataset, we question whether they can achieve comparable results on a more challenging benchmark that thoroughly exploits their reasoning abilities. In this paper, we introduce the COCO-MMR dataset, which increases the difficulty of the multi-modal reasoning task in three key aspects: larger data scale, more complex question forms, and broader question scope.

\begin{figure*}[ht]
\centering
\includegraphics[width=0.93\linewidth]{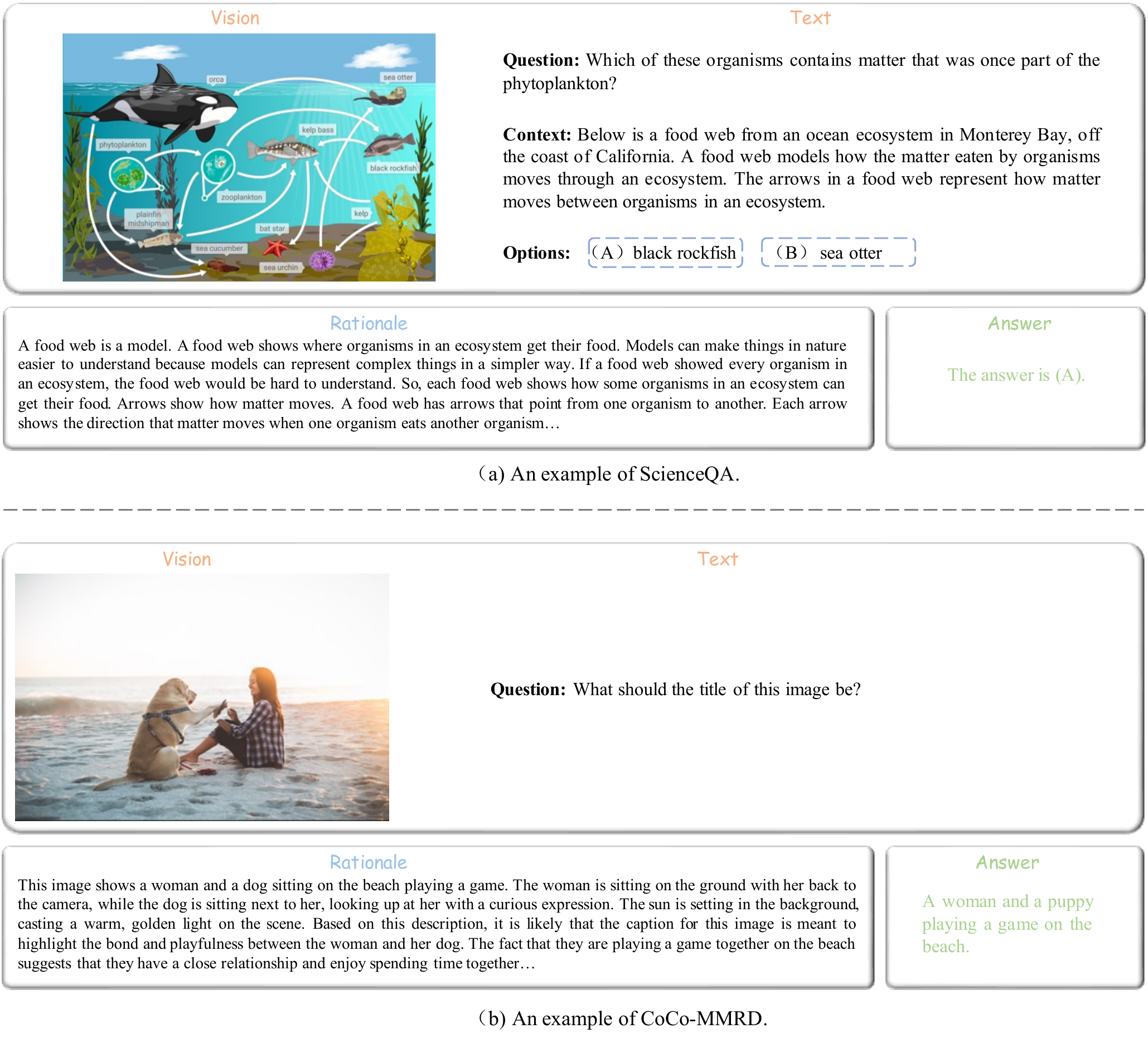}
\caption{The comparison of the ScienceQA dataset and the COCO-MMR dataset.}
\label{fig:dataset_comparison}
\end{figure*}

\section{COCO-MMR}

Our proposed COCO-MMR dataset differs from the ScienceQA dataset principally in the question forms and scope. As shown in Figure \ref{fig:dataset_comparison}, the ScienceQA dataset furnishes more granular information in the text and prompts the model to choose from provided options. In contrast, the COCO-MMR dataset demands that the model generate open-ended answers based solely on the question without additional context or options, posing a greater challenge. Furthermore, the COCO-MMR dataset encompasses a wider range of question types and scenarios, including those from daily life, which aligns more closely with real-world situations.

\subsection{Data Collection}

We illuminate the collection and processing methodology of our proposed COCO-MMR dataset here. The collected dataset derives from the COCO Caption dataset~\cite{chen2015microsoft} and the COCO VQA dataset~\cite{goyal2017making}.

\paragraph{Caption Dataset}

\begin{figure*}[htbp]
\centering
\includegraphics[width=0.96\linewidth]{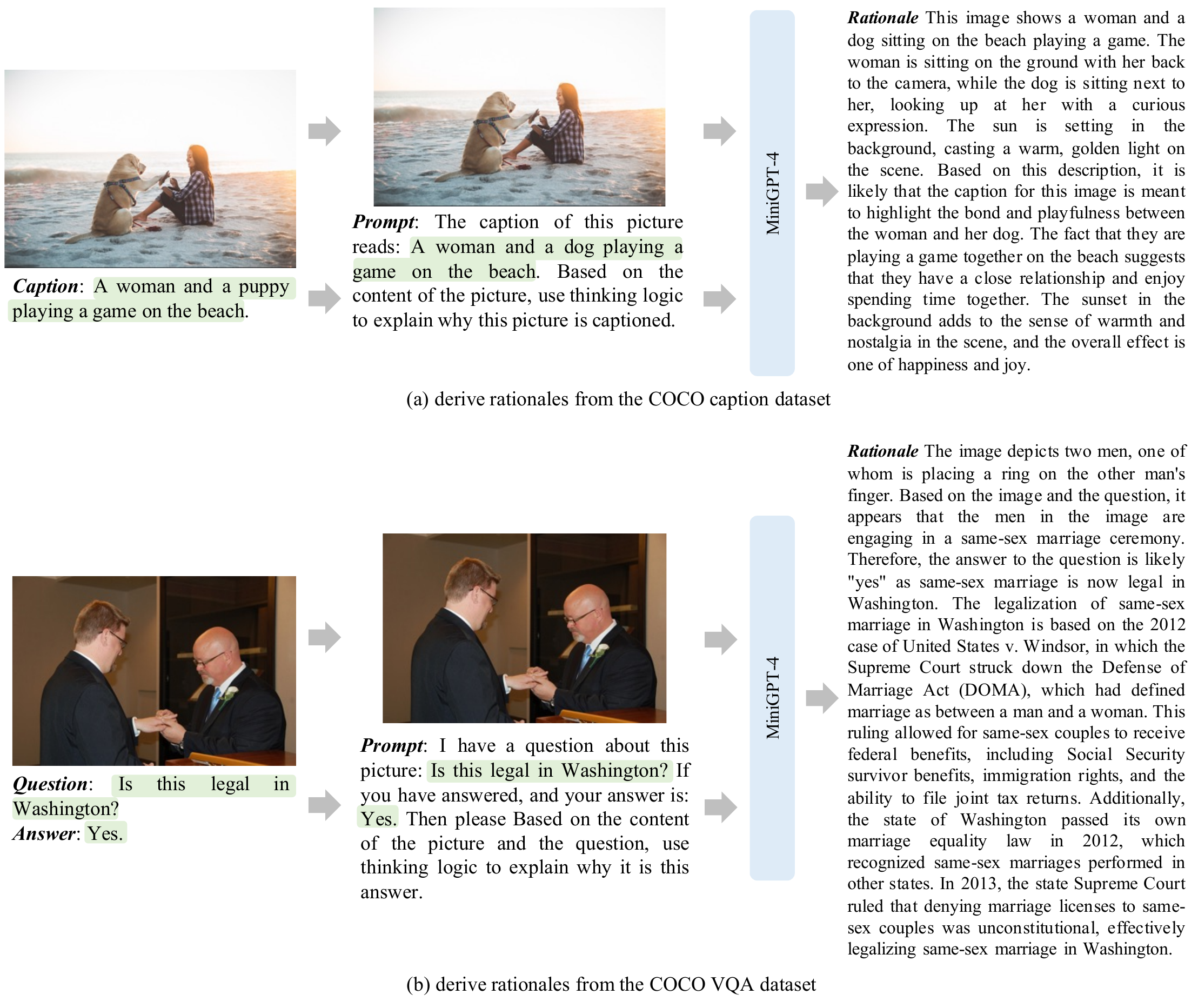}
\caption{The procedures of deriving rationales from (a) the COCO caption dataset and (b) the COCO VQA dataset.}
\label{fig:dataset_collection}
\end{figure*}

In the COCO caption dataset, the original captions from the COCO dataset were preserved and utilized as the foundation for developing cognitive reasoning inference processes, which were then combined with prompts. The main aim of these cognitive reasoning inferences was to generate answers that could be inferred as captions, using a combination of questions and images.

To facilitate the generation of thinking logic inferences by MiniGPT-4~\cite{zhu2023minigpt}, our data collection process employed a prompt template. As shown in Figure~\ref{fig:dataset_collection} (a), the template was structured as follows: "The caption of this picture reads: $[\texttt{Caption}]$. Based on the content of the picture, use thinking logic to explain why this picture is captioned as such." This prompt can effectively empower the MiniGPT-4 model to generate the desired and well-articulated rationales.

In total, we generated 30,000 triplets consisting of questions, rationales, and answers. Each data point underwent manual verification to ensure the quality and relevance of our dataset, confirming its alignment with the corresponding image.

\paragraph{VQA Dataset}

Similar to the COCO caption dataset, the COCO VQA dataset retained the original questions and generated the thinking logic inference process combined with prompts. The final objective was to generate rationales derived from a combination of questions and images.

As shown in Figure~\ref{fig:dataset_collection} (b), The data collection for the VQA dataset also utilized a prompt template: "I have a question about this picture: $[\texttt{Question}]$. If you have answered, and your ANSWER is: $[\texttt{Answer}]$. Then please, based on the content of the picture, the question and answer, use thinking logic to explain why it is this answer." This prompt guides the generation of informative rationales.

Initially, the VQA dataset contained 35,000 data points. After a meticulous manual filtering and detection process, we retained 32,351 data points that ensured the integrity of the dataset and the relevance of each data point to its corresponding image. The complete COCO-MMR dataset consists of 62,351 data points, which are divided into 56,115 training, 3,117 validation, and 3,119 test samples. To summarize, our data collection process was designed to create a dataset that encourages models to generate thoughtful and open-ended responses. The COCO-MMR dataset serves as a resource for evaluating models in multimodal reasoning.

\subsection{Dataset Analysis}
To procure a richer understanding of the dataset's structure, we've enumerated multiple features and executed a thorough study pertaining to each.

\begin{figure}[!t]
\begin{center}
\includegraphics[width=\linewidth]{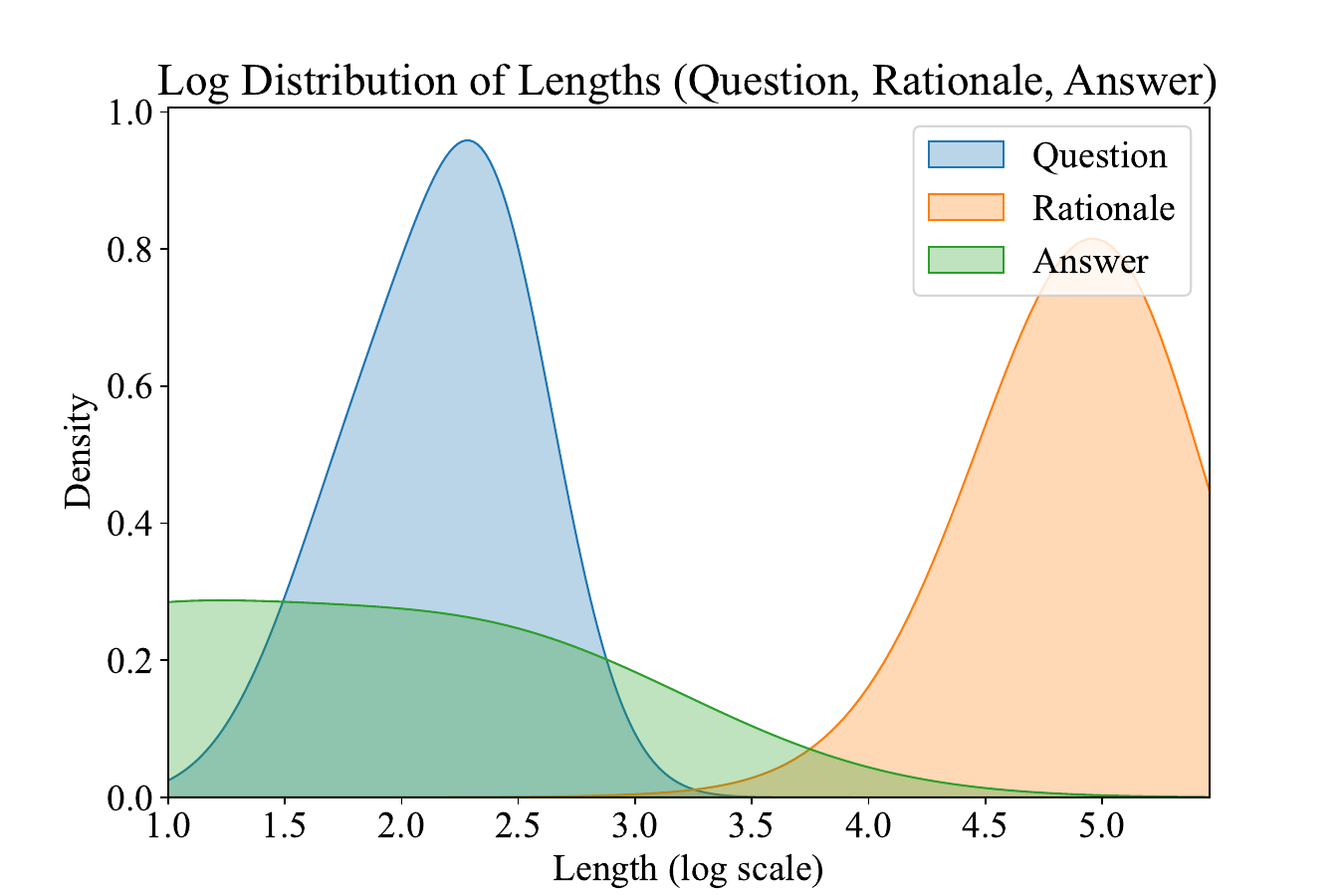}
\end{center}
\caption{The distribution of the lengths of questions, rationales, and answers in the dataset, where we take the log for improving visualization effects.}
% \vspace{-2mm}
\label{fig:question_length}
\end{figure}
\paragraph{Data Length}
We performed a statistical analysis on the character lengths of questions, rationales, and answers within the COCO-MMR dataset. As Figure \ref{fig:question_length} illustrates, the majority of question lengths in the dataset lie within a 2 to 10 character interval. Within this interval, the length of 10 characters frequently emerges, predominantly due to the consistent style of the title-related questions. The rationale character length exhibits a broader spread, varying between 50 to 250 characters. Lastly, the character lengths of answers predominantly cluster within a concise 1-20 character span.

\paragraph{Support Rate}
\begin{figure}[ht]
\begin{center}
\includegraphics[width=\linewidth]{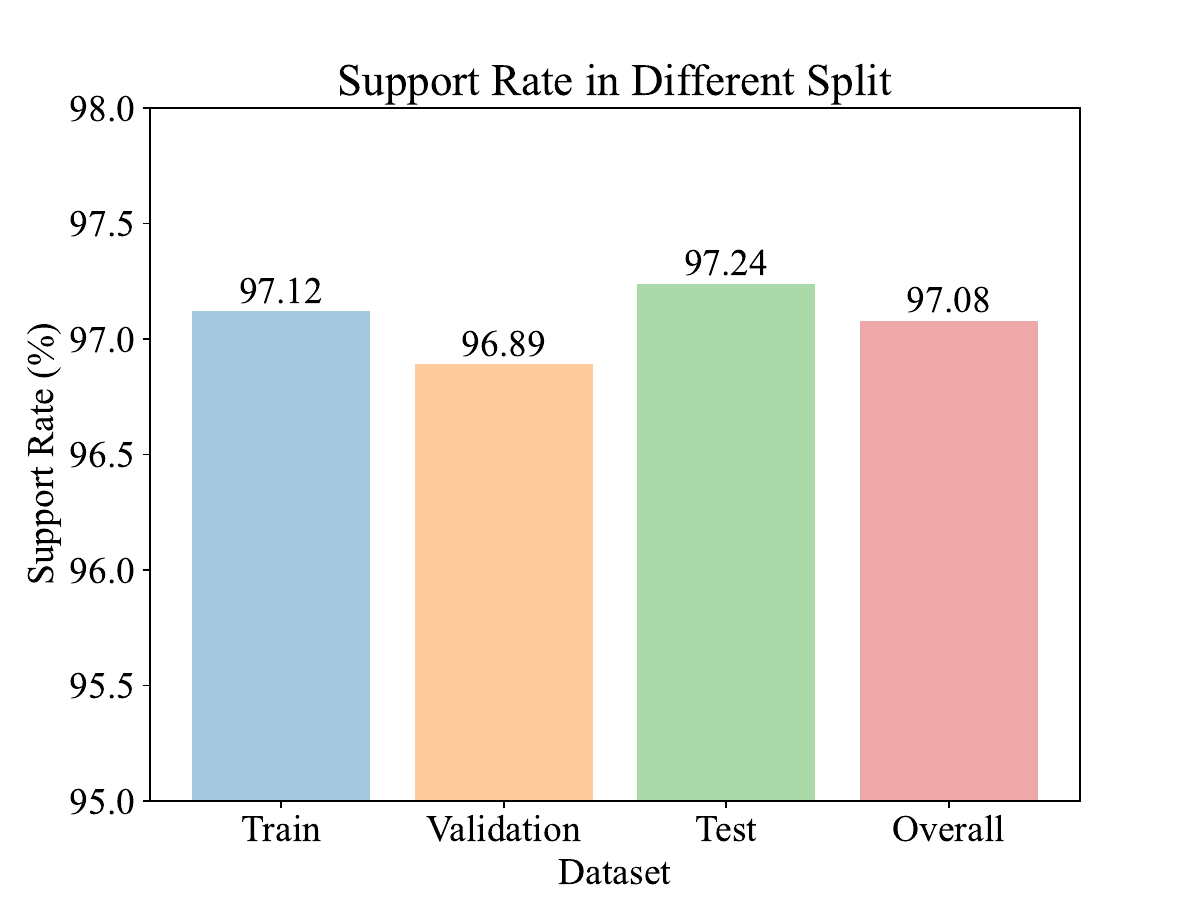}
\end{center}
\caption{Support Rate in Different Datasets.}
\vspace{-2mm}
\label{fig:Support_Rate}
\end{figure}
We introduce the concept of "Support Rate" as a measure of dataset reliability. For each triplet comprised of a question, rationale, and final answer, we evaluate whether the rationale and question adequately support the derivation of the final answer. If they are, it's categorized as "supportive" and conversely, as non-supportive. We combine the question and rationale as input and use a BERT-based textual entailment model to determine if the answer substantiates the hypothesis. Finally, we compute the proportion of supportive triplets (question, rationale, answer) in each dataset split. This proportion manifests as the support rate. The respective support rates for each split are portrayed in Figure~\ref{fig:Support_Rate}. As can be observed from the results, the majority of the rationales obtained through our data collection approaches are reasonable. This demonstrates that our COCO-MMR dataset is effective for the multimodal reasoning task.

\paragraph{Human Evaluation Analysis}

To conduct a human evaluation, we randomly chosed 200 data entries and assigned them to five volunteers for assessment. The primary evaluation criteria focused on determining if the reasoning could convincingly justify the derivation of the final answer from the given question. Interestingly, the results revealed that 92.20\% of the rationales effectively aided in deriving the final answer from the question. The disparity between this human evaluation and the support rate stems from the fact that the correct answers can be inferred from simple questions without relying on reliable rationales.

\begin{figure}[!t]
\vspace{-2mm}
\centering
\includegraphics[width=\linewidth]{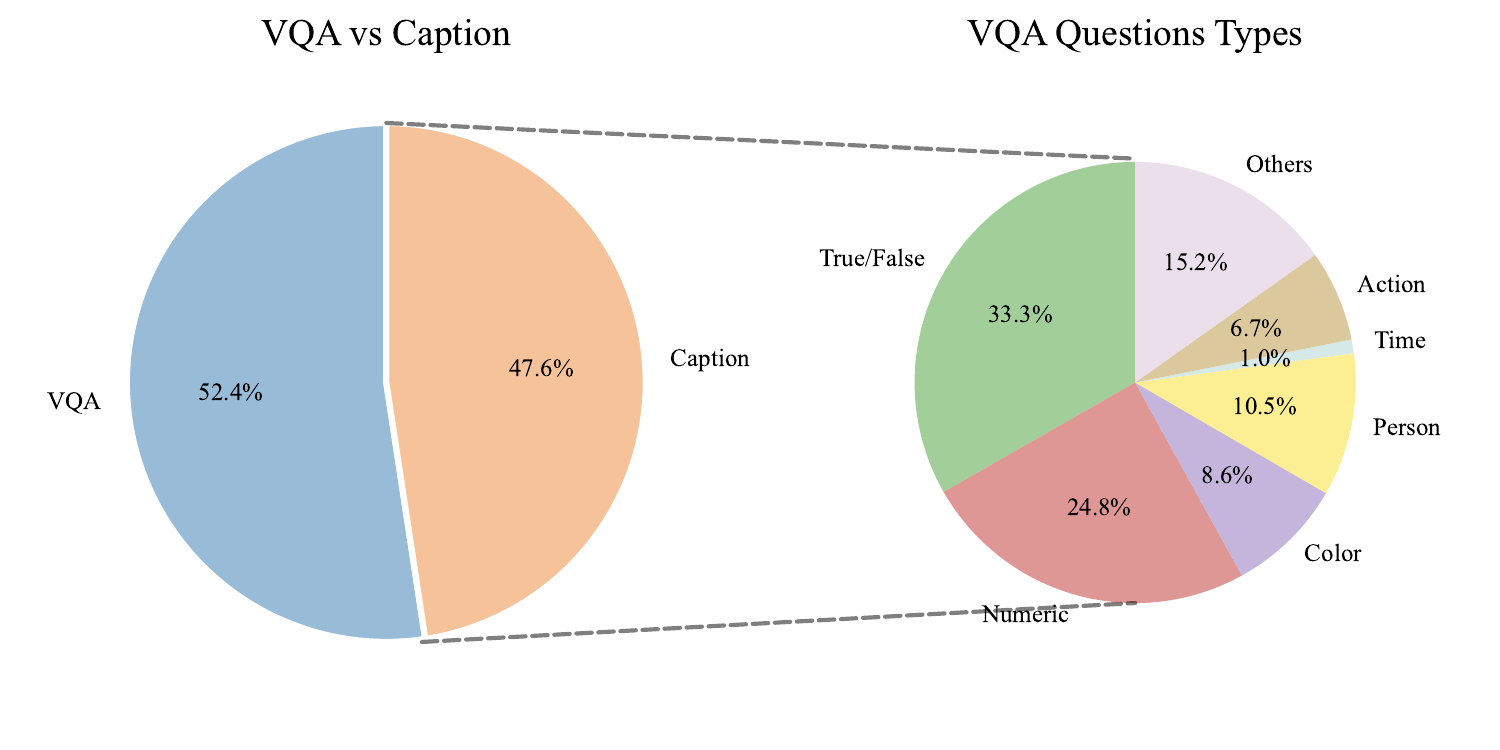}
\caption{The distribution of question types.}
\label{fig:Human_Evaluation}
\vspace{-2mm}
\end{figure}

In addition, the volunteers classified the selected 200 entries according to question type, with categories primarily including "Caption questions" and "VQA questions". The VQA questions were further divided into subcategories such as 'True/False Questions,' 'Numeric Questions,' 'Color Questions,' 'Person Questions,' 'Time Questions,' 'Action Questions,' and 'Others.' The distribution of these question types is illustrated in Figure~\ref{fig:Human_Evaluation}. It can be seen that the COCO-MMR dataset encompasses a wide variety of question types, enabling the extension of reasoning tasks to a broader range of real-world scenarios, thus posing a greater challenge for evaluation.

\section{Enigma-COT}

\begin{figure}[!t]
\centering
\includegraphics[width=0.98\linewidth]{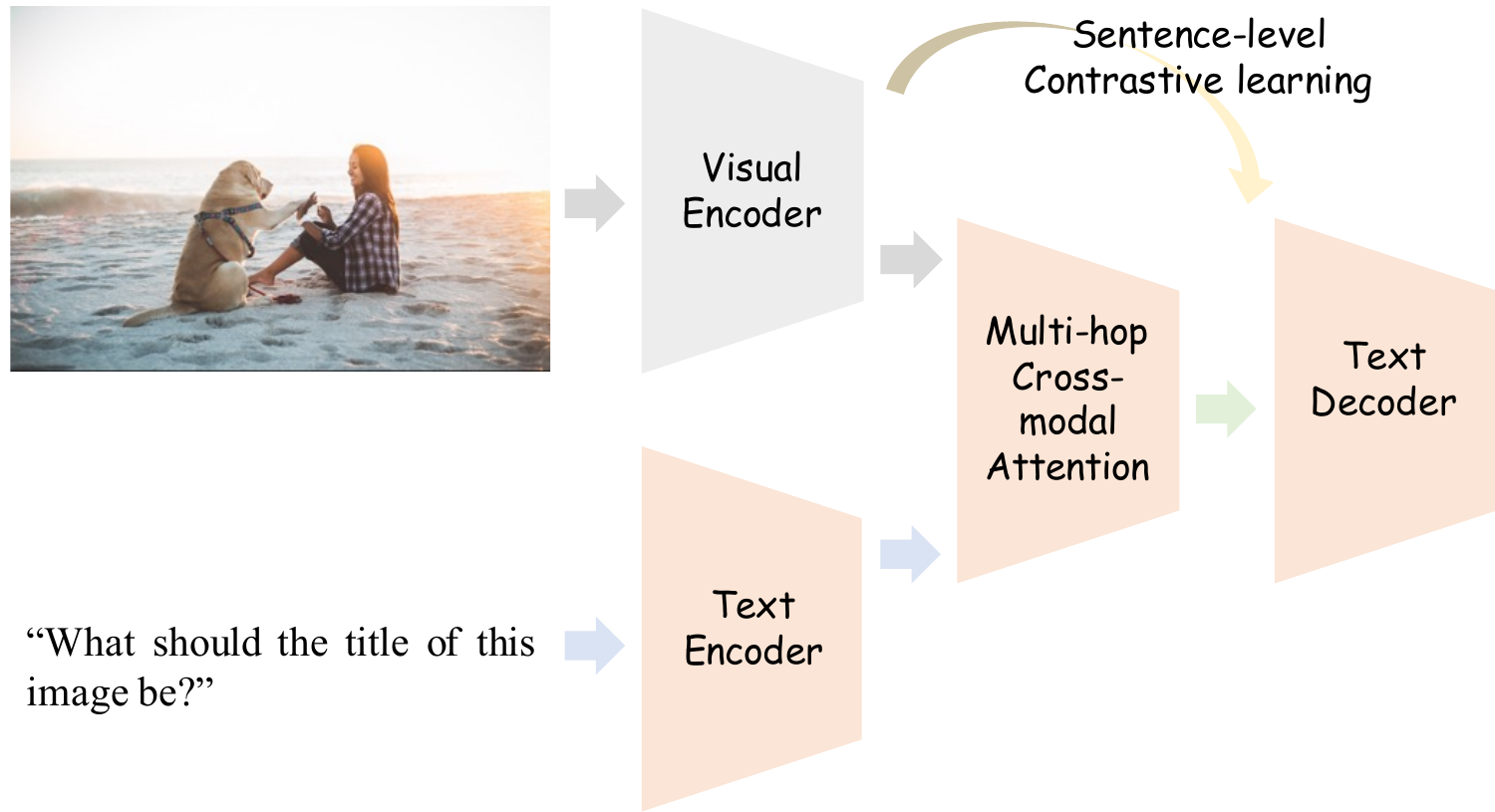}
\caption{The proposed model architecture. The gray visual encoder is frozen during training.}
\label{fig:model}
\end{figure}

We build a multimodal reasoning framework, Enigma-COT, to provide a strong baseline model for multimodal reasoning. Following~\cite{DBLP:journals/corr/abs-2302-00923}, our framework has two stages: rationale generation and answer inference. These stages use the same model architecture but have different inputs and outputs. In the first stage, we provide text and vision inputs to generate rationales. In the second stage, we combine the original text with the generated rationale. We then use the updated text and the original vision input to infer the answer. The model architecture is shown in Figure~\ref{fig:model}.

\subsection{Multi-Hop Cross-Modal Attention}

We propose multi-hop cross-modal attention that iteratively optimizes crucial information, resembling the repetitive thinking process of humans. The $k$-th hop attention $A^{k}$ is defined as follows:
\begin{equation}
A^{k} = \textrm{MHA}^{k}(H_t^{k-1}, H_i, H_i),
\end{equation}
where $\textrm{MHA}(\cdot)$ represents the multi-head attention layer~\cite{DBLP:conf/nips/VaswaniSPUJGKP17}. The $(k-1)$-th hop text hidden state $H_t^{k-1}$ is the query, while the image embedding $H_i$ acts as both the key and value.

The $k$-th hop text hidden state $H_t^{k}$ is obtained by aggregating the information of the previous hidden state $H_t^{k-1}$ and the attention $A^{k}$:
\begin{equation}
\begin{aligned}
H_t^{k} &= (1 - G^{k}) \odot H_t^{k-1} + G^{k} \odot A^{k}, \\
G^{k} &= \sigma(W^{k}[H_t^{(k-1)}, A^{k}]),
\end{aligned}
\end{equation}
where $\sigma$ denotes the sigmoid function, $\odot$ denotes element-wise multiplication, and $W^{(k)}$ is the $k$-th gating weight. The gating operation balances the text representation and the attention, enabling the model to better preserve the original information while effectively capturing and utilizing the key information captured by the current attention.

\subsection{Sentence-level Contrastive Learning}

While the multi-hop cross-modal attention is designed to enhance the fusion of visual and textual information, we also propose sentence-level contrastive learning to strengthen the text encoder. We maximize the similarity between the image embedding $H_i$ and the decoded text embedding $H_t$ at the sentence level, and control the similarity by a temperature parameter $\tau$:
\begin{equation}
  \mathrm{S}(H_i, H_t) = \frac{H_i \odot H_t^T}{\tau},
\end{equation}
and the sequence-level contrastive loss is:
\begin{equation}
  \mathcal{L}_{con} = -\log(\mathrm{softmax}(S(H_i, H_j))).
\end{equation}

The overall training objective is the linear combination of vanilla token-level cross-entropy loss and sentence-level contrastive loss:
\begin{equation}
  \mathcal{L} = \mathcal{L}_{ce} + \lambda \mathcal{L}_{con},
\end{equation}
where $\lambda$ is a parameter to balance the two losses.

\section{Experiments}

\begin{table*}[htbp]
  \small
  \centering
  \caption{Accuracy on the ScienceQA Dataset. Evaluation across eight categories: natural science, social science, linguistics, textual context, image context, no context, grades 1-6, and grades 7-12. The best results in each category are highlighted in bold.}
  {\renewcommand\baselinestretch{1.1}\selectfont
  \resizebox{0.86\textwidth}{!}{
  \begin{tabular}{l|c|cccccccc|c}
  \toprule
  Model                                   & \begin{tabular}[c]{@{}c@{}}Tuned\\ Params\end{tabular} & NAT              & SOC              & LAN              & TXT              & IMG              & NO               & G1-6             & G7-12            & AVG   \\
  \midrule
  Human~\cite{DBLP:conf/nips/LuMX0CZTCK22} & -     & 90.23 & 84.97 & 87.48 & 89.6  & 87.5  & 88.10  & 91.59 & 82.42 & 88.40 \\
  \hline
  MCAN~\cite{DBLP:conf/cvpr/Yu0CT019} & 95M   & 56.08 & 46.23 & 58.09 & 59.43 & 51.17 & 55.4  & 51.65 & 59.72 & 54.54 \\
  Top-Down~\cite{DBLP:conf/cvpr/00010BT0GZ18} & 70M   & 59.50  & 54.33 & 61.82 & 62.90  & 54.88 & 59.79 & 57.27 & 62.16 & 59.02 \\
  BAN~\cite{DBLP:journals/titb/YuPFCS23} & 112M  & 60.88 & 46.57 & 66.64 & 62.61 & 52.6  & 65.51 & 56.83 & 63.94 & 59.37 \\
  DFAF~\cite{DBLP:conf/cvpr/GaoJYLHWL19} & 74M   & 64.03 & 48.82 & 63.55 & 65.88 & 54.49 & 64.11 & 57.12 & 67.17 & 60.72 \\
  ViLT~\cite{DBLP:conf/icml/KimSK21} & 113M  & 60.48 & 63.89 & 60.27 & 63.2  & 61.38 & 57    & 60.72 & 61.90  & 61.14 \\
  Patch-TRM~\cite{lu2021iconqa} & 90M   & 65.19 & 46.79 & 65.55 & 66.96 & 55.28 & 64.95 & 58.04 & 67.50  & 61.42 \\
  VisualBERT~\cite{li2019visualbert,DBLP:conf/icml/KimSK21} & 111M  & 59.33 & 69.18 & 61.18 & 62.71 & 62.17 & 58.54 & 62.96 & 59.92 & 61.87 \\
  Mutimodal-CoT$_{\textrm{Base}}$~\cite{DBLP:journals/corr/abs-2302-00923} & 223M  & 87.52 & 77.17 & 85.82 & 87.88 & 82.90  & 86.83 & 84.65 & 85.37 & 84.91 \\
  Mutimodal-CoT$_{\textrm{Flan-T5-large}}$~\cite{DBLP:journals/corr/abs-2302-00923} & 783M  & \textbackslash{} & \textbackslash{} & \textbackslash{} & \textbackslash{} & \textbackslash{} & \textbackslash{} & \textbackslash{} & \textbackslash{} & 93.02 \\
  \hline
  UnifiedQA$_{\textrm{Base}}$~\cite{khashabi-etal-2020-unifiedqa} & 223M  & 68.16 & 69.18 & 74.91 & 63.78 & 61.38 & 77.84 & 72.98 & 65.00    & 70.12 \\
  UnifiedQA$_{\textrm{Base}}$ w/ CoT~\cite{DBLP:conf/nips/LuMX0CZTCK22} & 223M  & 71.00    & 76.04 & 78.91 & 66.42 & 66.53 & 81.81 & 77.06 & 68.82 & 74.11 \\
  \hline
  % LLaMA-Adapter$_{\textrm{T}}$ \cite{gao2023llamaadapterv2} & 1.2M  & 79.00    & 73.79 & 80.55 & 78.30  & 70.35 & 83.14 & 79.77 & 75.68 & 78.31 \\
  LLaMA-Adapter~\cite{gao2023llamaadapterv2} & 1.8M  & 84.37 & 88.30  & 84.36 & 83.72 & 80.32 & 86.90  & 85.83 & 84.05 & 85.19 \\
  LLaVa~\cite{DBLP:journals/corr/abs-2304-08485} & 13B  & 90.36 & 95.95  & 88.00 & 89.49 & 88.00 & 90.66  & 90.93 & 90.90 & 90.92 \\
  % LLaVa \cite{DBLP:journals/corr/abs-2304-08485} & 13B  & 91.56 & 96.74  & 91.09 & 90.62 & 88.99 & 93.52  & 92.73 & 92.16 & 92.53 \\
  \hline
\rowcolor{gray!30} Enigma-COT$_{\textrm{Base}}$ & 229M  & 88.28 & 78.74 & 85.64 & 88.51 & 84.28 & 86.90  & 85.43 & 85.89 & 85.59 \\
\rowcolor{gray!30} Enigma-COT$_{\textrm{Flan-T5-large}}$ & 793M  & \textbf{97.51} &  \textbf{84.70} &  \textbf{94.73} &  \textbf{96.68} &  \textbf{91.37} &  \textbf{95.89} &  \textbf{94.46} &  \textbf{93.47} & \textbf{94.11} \\
  \bottomrule
  \end{tabular}}\par}
\label{tab:addlabel1}
\end{table*}

In this section, we provide detailed information about the experiments conducted on the ScienceQA and COCO-MMR.

\subsection{Experimental Setup}
Enigma-COT is initialized by utilizing UnifiedQA \cite{khashabi-etal-2020-unifiedqa} and FLAN-T5 \cite{DBLP:journals/corr/abs-2210-11416}. We incorporate an early stopping strategy to stop the training when the validation loss achieves stability. The maximum number of training epochs is $50$, with a batch size of $32$ and a learning rate of $5\times 10^{-5}$. All experiments were conducted using four NVIDIA Tesla A100 80G GPUs.

% For the ScienceQA dataset, we configure both the rationale generation and answer generation stages with a maximum input sequence length of 512, while the output lengths are defined as 512 and 64 respectively. When it comes to the COCO-MMR dataset, the rationale generation and answer generation stages are regulated with a maximum input sequence length of 100 and 400 respectively, and the output lengths are set to 300 and 100, correspondingly. All experiments were conducted using four NVIDIA Tesla A100 80G GPUs.

\subsection{Baseline Models}
\paragraph{For the ScienceQA dataset}
In alignment with \cite{DBLP:journals/corr/abs-2302-00923,gao2023llamaadapterv2}, the baseline models encompass: (i) human evaluation \cite{DBLP:conf/nips/LuMX0CZTCK22}, (ii) visual question answering models \cite{DBLP:conf/cvpr/Yu0CT019,DBLP:conf/cvpr/00010BT0GZ18,DBLP:journals/titb/YuPFCS23,DBLP:conf/cvpr/GaoJYLHWL19,DBLP:conf/icml/KimSK21,lu2021iconqa,li2019visualbert}, (iii) text-to-text generation (LM) models \cite{khashabi-etal-2020-unifiedqa,DBLP:conf/nips/LuMX0CZTCK22}, and (iv) Multimodal LLM models \cite{gao2023llamaadapterv2,DBLP:journals/corr/abs-2302-00923}. The performance of all baseline models is encapsulated in Table \ref{tab:addlabel1}.

\paragraph{For the COCO-MMR dataset} We select several representative approaches above to evaluate them on the COCO-MMR dataset, including: (i) human evaluation, (ii) visual question answering models~\cite{DBLP:journals/corr/abs-2302-00923}, (iii) text-to-text generation (LM) models~\cite{khashabi-etal-2020-unifiedqa,DBLP:conf/nips/LuMX0CZTCK22}, and (iv) Multimodal LLM models \cite{DBLP:journals/corr/abs-2304-08485}. 

For the methods of human evaluation, we recruited 20 volunteers (with educational backgrounds ranging from bachelor's to doctoral degrees). Each volunteer was randomly assigned 50 questions from the complete test set to answer. We created a straightforward reading webpage for humans to read the questions and images and write their responses. Participants only needed to write the answers based on the question and corresponding image. Monetary rewards were given upon test completion.

% For the detailed evaluation metrics of these models, please refer to Appendix~\ref{metrics}.
\subsection{Evaluation Metrics}
\label{metrics}
For the evaluation of our multimodal reasoning chain task, we utilize four main metrics: AVG Accuracy, BLEU, Similarity, and Rouge. AVG Accuracy serves as the ultimate evaluation criterion for the ScienceQA dataset, while Rouge is the primary metric for the rationale generation phase of the ScienceQA dataset and for the COCO-MMR dataset as well.

\textbf{Accuracy}
We apply accuracy on the ScienceQA dataset to measure the model's precision in predicting answers after generating rationales. The dataset spans eight aspects: natural science, social science, linguistics, textual context, image context, no context, and for grade levels 1-6 and 7-12. We calculated the accuracy for each of these aspects and their average accuracy. The accuracy for each aspect is determined by the ratio of the number of correctly predicted answers for questions within that aspect over the total number of questions predicted by the model within that aspect, represented by the equation:

\begin{equation}
\text{Accuracy} = \frac{\text{Number of Correctly Predicted Samples}}{\text{Total Number of Predicted Samples}}
\end{equation}

The average accuracy is the mean of all these individual accuracies, denoted by the following equation:

\begin{equation}
\text{Average Accuracy} = \frac{1}{n}\sum_{i=1}^{n} \text{Accuracy}_i
\end{equation}

Here, $n$ refers to the number of aspects (in this instance, $n=8$), and $\text{Accuracy}_i$ is the accuracy of the $i$th aspect.

\textbf{BLEU (Bilingual Evaluation Understudy)}: BLEU is an automatic evaluation metric based on n-gram precision, originally used for machine translation tasks. In our task, BLEU mainly evaluates the lexical overlap between the generated rationale or answer and the reference text. Given a reference text $r$ and a candidate text $c$, BLEU is defined as follows:

\begin{equation}
\text{BLEU} = \text{BP} \cdot \exp \left( \sum_{n=1}^{N} w_n \log p_n \right)
\end{equation}

where $p_n$ is the $n$-gram precision, $N$ is the maximum order of n-gram, and $w_n$ is the weight of the n-gram precision, usually set to $1/N$. BP (Brevity Penalty) is used to penalize too short candidate texts and is defined as follows:

\begin{equation}
\text{BP} = \left\{ \begin{array}{ll}
1 & \mbox{if } c > r \\
\exp(1 - r / c) & \mbox{otherwise}
\end{array} \right.
\end{equation}

\textbf{Similarity}: Similarity is mainly used to evaluate the semantic similarity of the generated rationale or answer. In our task, we use cosine similarity as the similarity metric, defined as follows:

\begin{equation}
\text{Similarity} = \frac{\textbf{c} \cdot \textbf{r}}{||\textbf{c}||_2 ||\textbf{r}||_2}
\end{equation}

where $\textbf{c}$ and $\textbf{r}$ represent the embedding vectors of the candidate text and reference text, respectively.

\textbf{ROUGE (Recall-Oriented Understudy for Gisting Evaluation)}: ROUGE is an evaluation metric widely used in automatic summarization and machine translation tasks, mainly used to measure the recall performance of the generated text and the reference text. In our task, we mainly use the ROUGE-L metric, which evaluates the similarity between texts based on the longest common subsequence (LCS). ROUGE-L is defined as follows:

\begin{equation}
\text{ROUGE-L} = \frac{\text{LCS}(c, r)}{\max(\text{length}(c), \text{length}(r))}
\end{equation}

where $\text{LCS}(c, r)$ is the length of the longest common subsequence between the candidate text $c$ and the reference text $r$.

\subsection{Experimental Results}

Table~\ref{tab:addlabel1} presents the comparative experimental results on the ScienceQA dataset. The results demonstrate that humans achieved an average accuracy rate of 88.40\%. Notably, both the Mutimodal-CoT and LLaVa models surpassed human performance, achieving accuracy rates of 93.02\% and 90.92\% respectively. Remarkably, the Enigma-COT model achieved an average precision of 94.11\%, representing a significant breakthrough on this benchmark dataset. These findings emphasize the advanced capabilities of our proposed multimodal model, Enigma-COT, in effectively addressing the intricacies of multimodal reasoning by employing the innovative chain-of-thought technique.

Furthermore, the aforementioned research results clearly indicate that the ScienceQA dataset, with its limited scale, reliance on multiple-choice questions, and constraints in scientific reasoning, is not adequately prepared to embrace future advancements in multimodal reasoning. Therefore, there is a pressing need to introduce a more complex dataset.

\begin{table*}[htbp]
  \small
  \centering
  \caption{The performance comparison on the COCO-MMR Dataset.}
  {\renewcommand\baselinestretch{1.3}
%   \selectfont\resizebox{\textwidth}{!}{
  \begin{tabular}{lccccc}
  \toprule
  Model & Type & Tuned Params & BLUE & Similarity & Rouge \\ 
  \midrule
  Human & Human & - & 80.35 & 90.25 & 82.73 \\ 
  \hline
  Mutimodal-CoT$_{\textrm{Flan-T5-Large}}$~\cite{DBLP:journals/corr/abs-2302-00923} & VQA & 783M &  66.89 & 79.77 & 69.04 \\ 
  \hline
  UnifiedQA$_{\textrm{Base}}$~\cite{khashabi-etal-2020-unifiedqa} & Text-to-text  & 223M & 43.76 & 63.97 & 39.46 \\
  UnifiedQA$_{\textrm{Base}}$ w/ COT~\cite{DBLP:conf/nips/LuMX0CZTCK22} & Text-to-text & 223M & 64.71 & 75.75 & 66.62 \\ 
  \hline
  LLAVA \cite{DBLP:journals/corr/abs-2304-08485}            & LLM & 13B & 59.38 & 74.59 & 62.15 \\ 
  \hline
  \rowcolor{gray!30}Enigma-COT$_{\textrm{Base}}$ & Ours & 229M & 65.52 & 78.03 & 67.67 \\
  \rowcolor{gray!30}Enigma-COT$_{\textrm{Flan-T5-large}}$ & Ours & 793M & \textbf{68.07} & \textbf{80.41} & \textbf{70.17} \\ 
  \bottomrule
  \end{tabular}
%   \par}
  }
  \label{tab:addlabel5}
  \end{table*}

\begin{figure}[!t]
\begin{center}
\includegraphics[width=\linewidth]{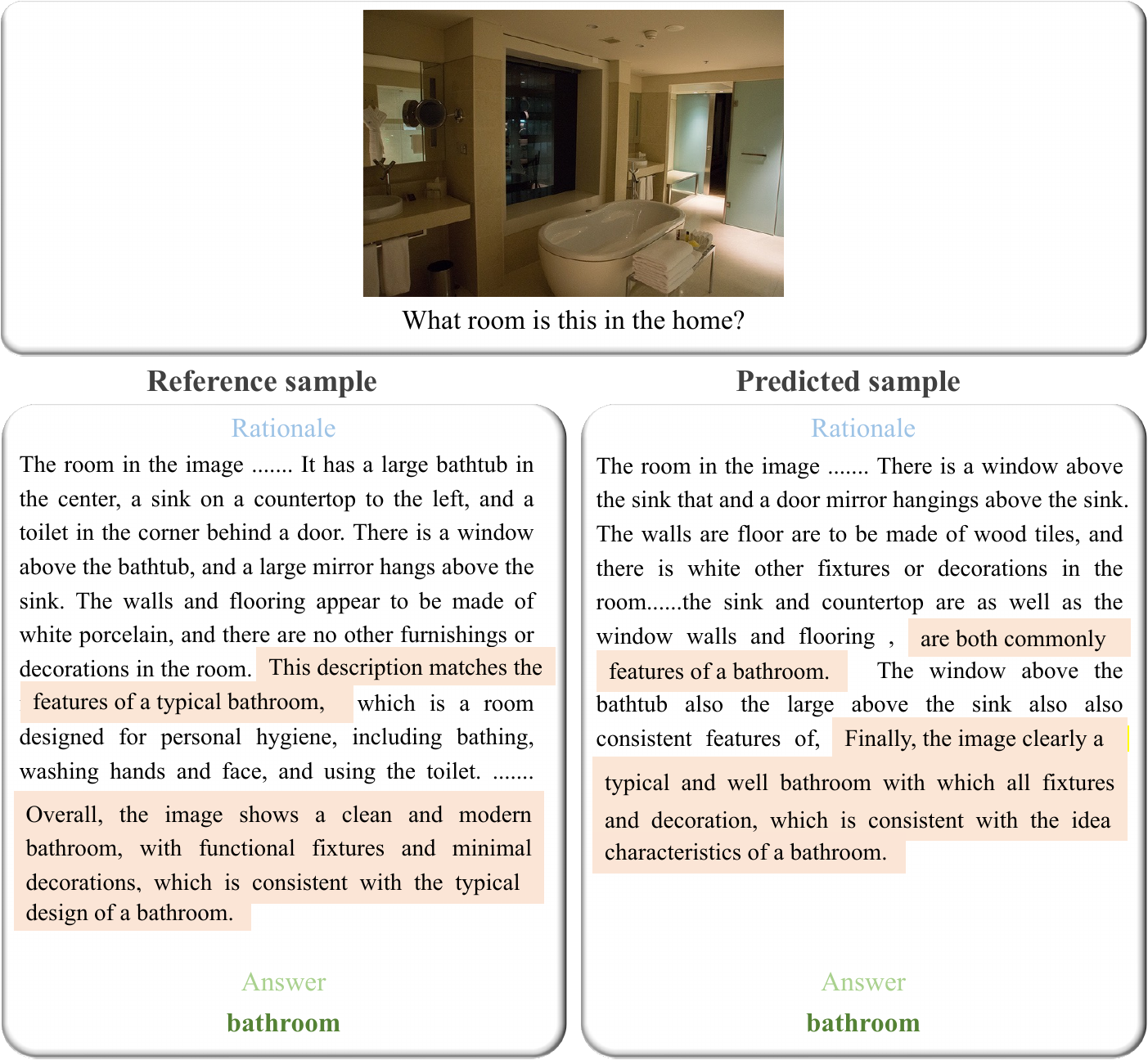}
\end{center}
\caption{An example of correct predictions using Enigma-COT on the COCO-MMR dataset.}
\label{fig:example1}
\end{figure}

Table~\ref{tab:addlabel5} provides a performance comparison among the representative approaches on the COCO-MMR dataset. Following~\cite{DBLP:journals/corr/abs-2302-00923}, the principal evaluation metrics encompass BLUE, Similarity, and Rouge scores. Within the COCO-MMR dataset, the Rouge score for humans is recorded as 82.73\%. The scores for Multi-modal-CoT, UnifiedQA, LLaVa, and Enigma-COT are 69.04\%, 66.62\%, 62.15\%, and 70.17\% respectively. Even though Enigma-COT attains the top performance, a substantial discrepancy still persists between its results and human performance. This suggests that despite some progress made by machine learning methods on this inference tasks, the complex cognition and understanding abilities of humans remain inimitable. This presents new challenges for future research. There is a need to further optimize our models in order to lessen the performance gap between human and machine on this particular task.
Figure~\ref{fig:example1} displays an example of the predictions made by the Enigma-COT model on the COCO-MMR dataset. As depicted, the dataset incorporates open-ended questions, requiring the model to comprehend the image and the question, and, leveraging common-sense knowledge, establish connections between the rooms in the image and various locations within a home. Finally, based on the inferred results, the model generates an answer indicating the specific location in the home represented by the image.
% In addition, a detailed analysis of further prediction results is carried out in Appendix~\ref{Example_Analysis}. For the analysis of Enigma-COT in the rational generation, please refer to Appendix~\ref{Experimental_Results_2}.

\subsection{Ablation Study}
Table \ref{tab:addlabel2} represents the ablation study conducted on the ScienceQA dataset. Using Enigma-COT$_{\textrm{Base}}$ as the base model, the removal of the Multi-Hop Cross-Modal Attention module resulted in a more pronounced performance drop compared to removing the Sentence-level Contrastive Learning module.
\begin{table}[htbp]
  \small
  \centering
  \caption{Ablation study on the ScienceQA.}
  {\renewcommand\baselinestretch{1.2}\selectfont
  \resizebox{\linewidth}{!}{
    \begin{tabular}{l|ccccccccc}
    \toprule
    Model & NAT & LAN & TXT & IMG & G1-6 & AVG \\
    \midrule
    Enigma-COT$_{\textrm{Base}}$ & 87.97 & 85.91 & 88.27 & 84.09 & 85.24 & 85.33 \\
    w/o CL & 87.52 & 86.73 & 87.78 & 83.54 & 84.95 & 85.26 \\
    w/o Multi-hop & 87.61 & 86.00 & 87.83 & 83.44 & 84.91 & 85.17 \\
    \bottomrule
    \end{tabular}\par}}%
    % }
  \label{tab:addlabel2}%
\end{table}%
Table \ref{tab:addlabel7} represents the ablation study conducted on the COCO-MMR dataset. Similarly, it can be observed that the Multi-Hop Cross-Modal Attention module is more crucial for improving the model's performance.
    \begin{table}[htbp]
        \centering
        \caption{Ablation study on the COCO-MMR.}
        {\renewcommand\baselinestretch{1.2}\selectfont
  \resizebox{\linewidth}{!}{
        \begin{tabular}{lcccc}
          \toprule
          Model  & Tuned Params & BLUE  &  Similarity & Rouge \\ 
          \midrule
          
          Enigma-COT$_{\textrm{Base}}$                                                                                                                              & 229M                                                            & 65.52          & 78.03                                      & 67.67                                 \\
          % $_{\textrm{w/o CL}}$            
          w/o CL & 229M   & 65.49          & 77.96                                      & 67.58                                 \\ 
          w/o Multi-hop                                                         & 223M                                                            & 65.38          & 77.03                                      & 67.41                                 \\
          \bottomrule
          \end{tabular}
      }\par}
    \label{tab:addlabel7}%
  \end{table}%

% \subsection{Case Analysis of the ScienceQA Dataset}

\subsection{Case Analysis}
\label{Example_Analysis}
To gain a deeper understanding of the performance of the Enigma-COT model and the nature of the COCO-MMR dataset, we manually selected and analyzed a random sample of 500 data entries.

\begin{figure}[htbp]
\begin{center}
\includegraphics[width=\linewidth]{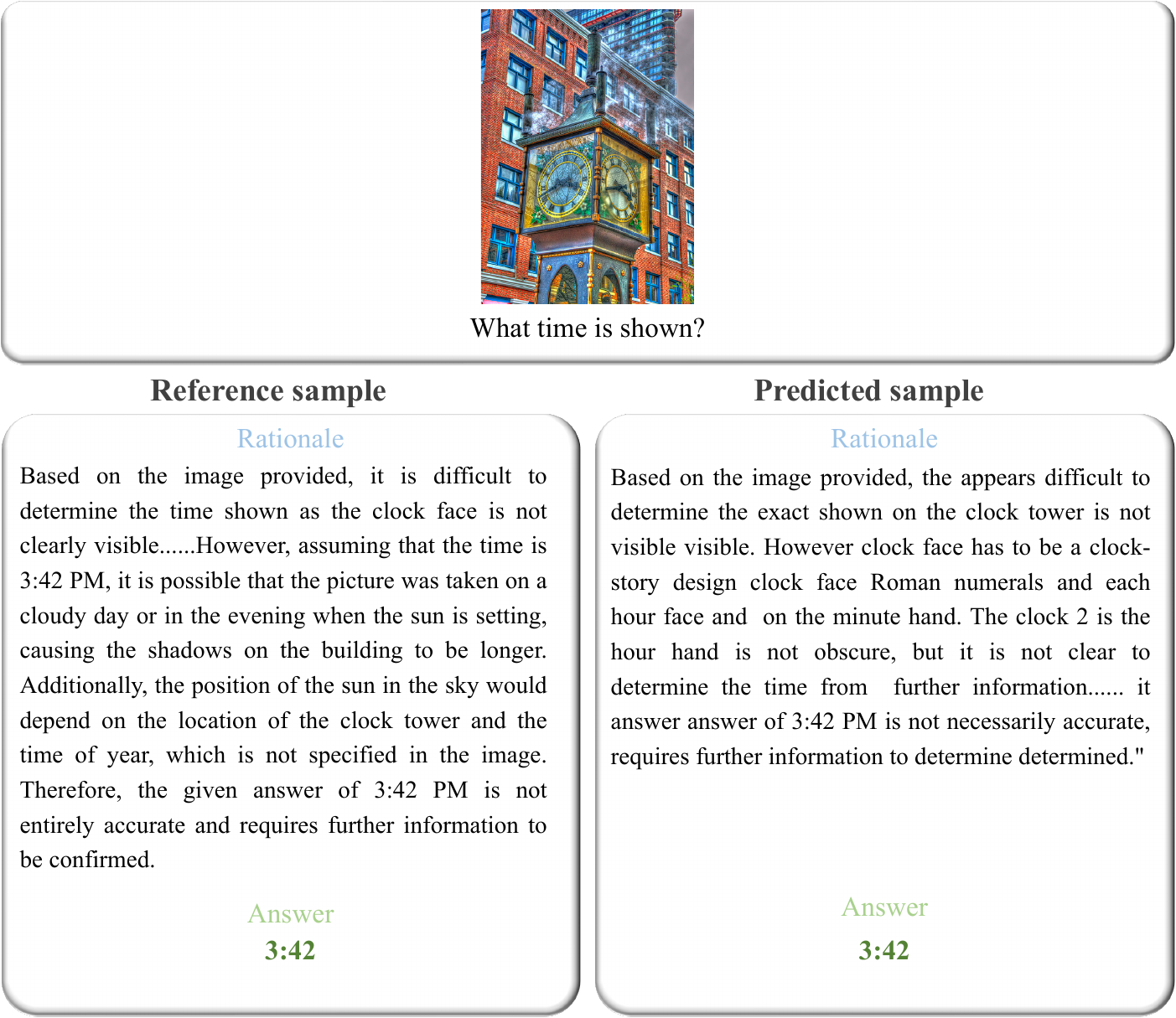}
\end{center}
\caption{Examples of Cases where Original Rationale is Invalid, Resulting in Invalid Reasoning Stage Rationale but Correct Predictions on the COCO-MMR Dataset.}
\label{fig:example2}
\end{figure}

\begin{figure}[htbp]
\begin{center}
\includegraphics[width=\linewidth]{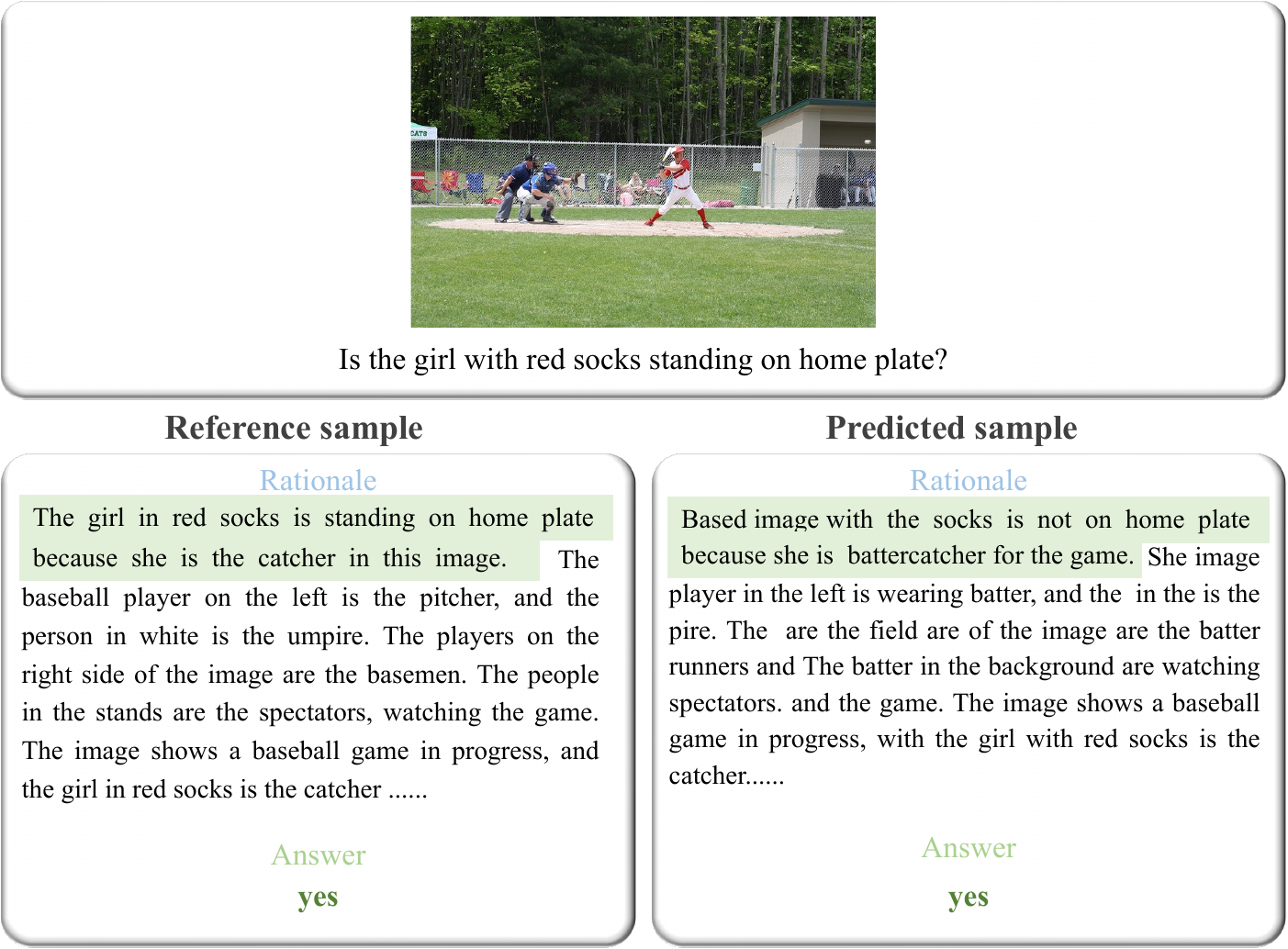}
\end{center}
\caption{Examples of Cases where Original Rationale is Correct, Reasoning Stage Rationale is Incorrect, but Predictions are Correct on the COCO-MMR Dataset.}
\label{fig:example3}
\end{figure}

We found that samples with a correctly predicted Answer do not necessarily contain a valid Rationale. In Figure \ref{fig:example2}, the original Rationale was invalid, resulting in an invalid Rationale during the reasoning stage, yet the Answer was correct. Moreover, another situation emerged, as shown in Figure \ref{fig:example3}, where the original Rationale was correct, but the Rationale during the reasoning stage was incorrect, yet the predicted Answer was correct. These results suggest that the Chain of Thought (COT) might not be useful for inferring the Answer in some data instances and can be ignored.

\begin{figure}[htbp]
\begin{center}
\includegraphics[width=\linewidth]{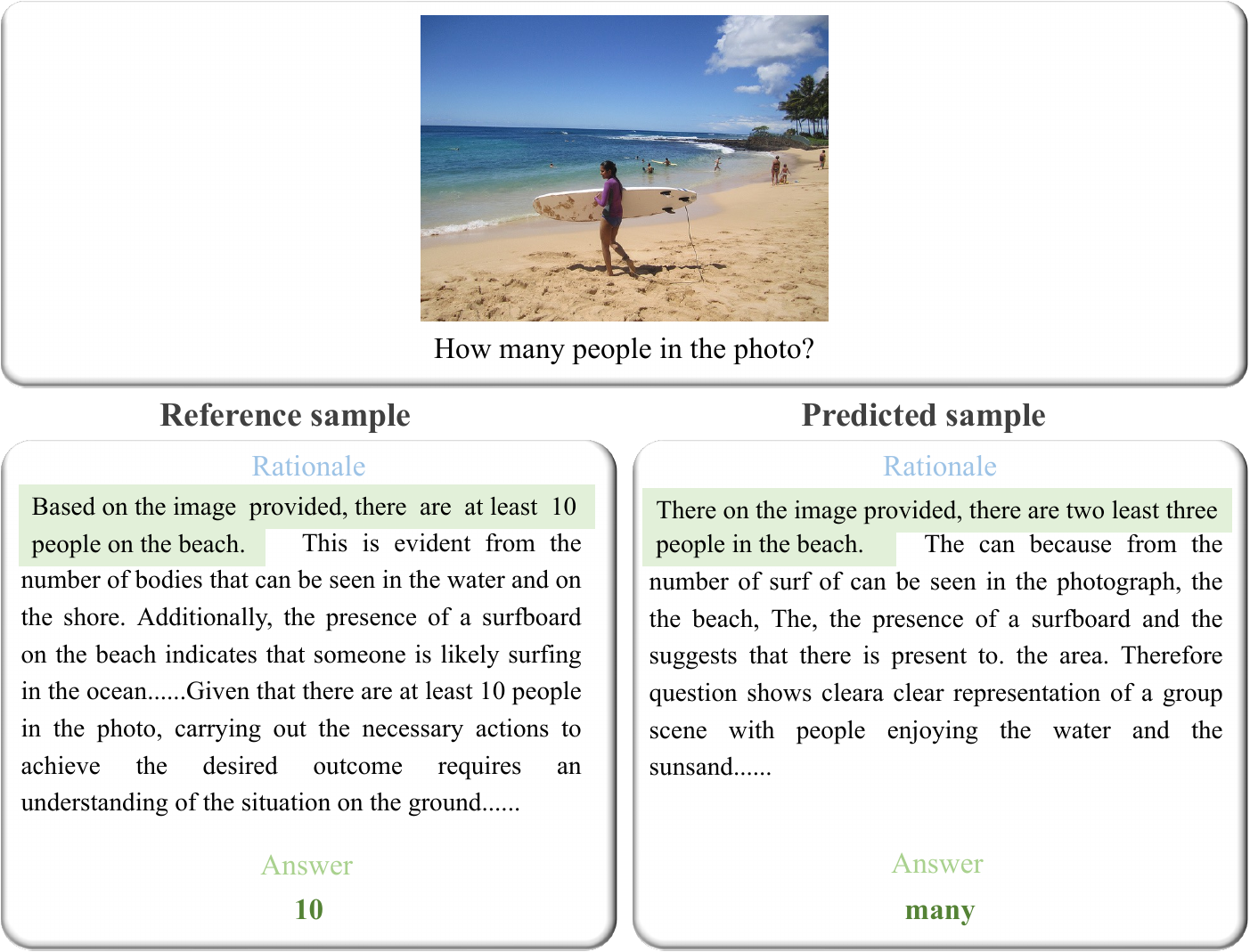}
\end{center}
\caption{Examples of Cases where Original Rationale and Reasoning Stage Rationale are Correct but Predictions are Incorrect on the COCO-MMR Dataset.}
\label{fig:example4}
\end{figure}

\begin{figure}[htbp]
\begin{center}
\includegraphics[width=\linewidth]{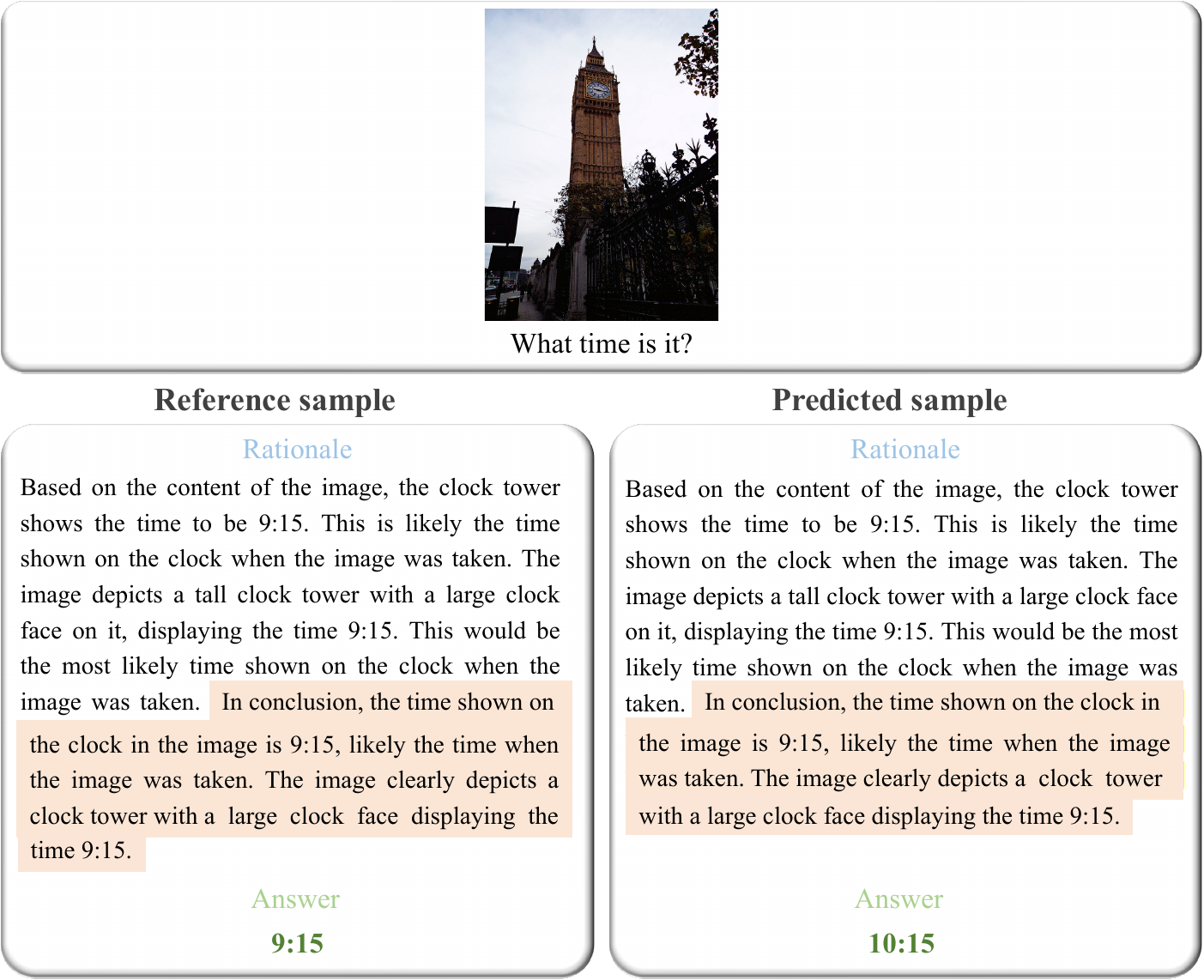}
\end{center}
\caption{Examples of Cases where Original Rationale is Correct, Reasoning Stage Rationale is Incorrect, and Predictions are also Incorrect on the COCO-MMR Dataset.}
\label{fig:example5}
\end{figure}

We also found that in samples where the predicted Answer was incorrect, the original Rationale was correct, and the Rationale during the reasoning stage was incorrect, resulting in an incorrect predicted Answer, as shown in Figure \ref{fig:example4}. Additionally, another scenario presented itself where the original Rationale was correct, the Rationale during the reasoning stage was correct, but the predicted Answer was incorrect, as depicted in Figure \ref{fig:example5}.

\subsection{Applications and Potential Impacts of the Dataset}
\label{Applications_and_Potential}
From the perspective of application scenarios, COCO-MMR, as a multimodal reasoning dataset, has a wide range of uses. It can be employed in general domains to train models to understand and generate more complex and diversified responses. This not only promotes the complexity and diversity of AI but also propels its application in real-world scenarios. Moreover, the COCO-MMR dataset can also be used to improve the interpretability of models, allowing us to comprehend how models infer answers from questions and images. This plays a vital role in enhancing the reliability and trustworthiness of models.

Regarding potential impacts, the COCO-MMR dataset may profoundly influence the development of AI. Specifically, it can assist us in developing superior dialogue systems, search engines, and even intelligent assistants, which will significantly increase the efficiency of our daily lives and work. However, we must be vigilant about the serious problems that may arise from the misuse of these technologies. For instance, some may exploit these technologies to fabricate fake news or conduct phishing on the internet. Therefore, as we advance, we also need to be alert to prevent its potential adverse impacts and proactively take measures to mitigate them.

\section{Conclusion}
In addressing the multimodal reasoning chain problem, our focus is on enabling models to utilize common sense knowledge and reasoning in a manner akin to humans, thereby understanding a broad array of concepts and situations encountered in daily life. We introduce a challenging multimodal reasoning dataset, COCO-MMR, that incorporates an open-ended question format, along with a pioneering model named Enigma-COT. Our experimental results demonstrate the effectiveness of our proposed dataset and model. We anticipate that our efforts will further the advancement of models boasting robust reasoning abilities applicable to diverse real-world scenarios and catalyze future exploration in multimodal reasoning.

\section*{Acknowledgments}
This work was supported by the National Key R\&D Program of China (2022ZD0115100), the National Natural Science Foundation of China (U21A20427), the Competitive Research Fund (WU2022A009) from the Westlake Center for Synthetic Biology and Integrated Bioengineering, and the Liaoning Province Applied Basic Research Program (20220303).

\bibliographystyle{IEEEtran}
\bibliography{bare_jrnl_new_sample4}

\begin{thebibliography}{10}
\providecommand{\url}[1]{#1}
\csname url@samestyle\endcsname
\providecommand{\newblock}{\relax}
\providecommand{\bibinfo}[2]{#2}
\providecommand{\BIBentrySTDinterwordspacing}{\spaceskip=0pt\relax}
\providecommand{\BIBentryALTinterwordstretchfactor}{4}
\providecommand{\BIBentryALTinterwordspacing}{\spaceskip=\fontdimen2\font plus
\BIBentryALTinterwordstretchfactor\fontdimen3\font minus
  \fontdimen4\font\relax}
\providecommand{\BIBforeignlanguage}[2]{{%
\expandafter\ifx\csname l@#1\endcsname\relax
\typeout{** WARNING: IEEEtran.bst: No hyphenation pattern has been}%
\typeout{** loaded for the language `#1'. Using the pattern for}%
\typeout{** the default language instead.}%
\else
\language=\csname l@#1\endcsname
\fi
#2}}
\providecommand{\BIBdecl}{\relax}
\BIBdecl

\bibitem{DBLP:conf/nips/2022}
\BIBentryALTinterwordspacing
S.~Koyejo, S.~Mohamed, A.~Agarwal, D.~Belgrave, K.~Cho, and A.~Oh, Eds.,
  \emph{Advances in Neural Information Processing Systems 35: Annual Conference
  on Neural Information Processing Systems 2022, NeurIPS 2022, New Orleans, LA,
  USA, November 28 - December 9, 2022}, 2022. [Online]. Available:
  \url{https://papers.nips.cc/paper_files/paper/2022}
\BIBentrySTDinterwordspacing

\bibitem{tan2022hyperspherical}
C.~Tan, Z.~Gao, L.~Wu, S.~Li, and S.~Z. Li, ``Hyperspherical consistency
  regularization,'' in \emph{Proceedings of the IEEE/CVF Conference on Computer
  Vision and Pattern Recognition}, 2022, pp. 7244--7255.

\bibitem{DBLP:journals/corr/abs-2201-08239}
\BIBentryALTinterwordspacing
R.~Thoppilan, D.~D. Freitas, J.~Hall, N.~Shazeer, A.~Kulshreshtha, H.~Cheng,
  A.~Jin, T.~Bos, L.~Baker, Y.~Du, Y.~Li, H.~Lee, H.~S. Zheng, A.~Ghafouri,
  M.~Menegali, Y.~Huang, M.~Krikun, D.~Lepikhin, J.~Qin, D.~Chen, Y.~Xu,
  Z.~Chen, A.~Roberts, M.~Bosma, Y.~Zhou, C.~Chang, I.~Krivokon, W.~Rusch,
  M.~Pickett, K.~S. Meier{-}Hellstern, M.~R. Morris, T.~Doshi, R.~D. Santos,
  T.~Duke, J.~Soraker, B.~Zevenbergen, V.~Prabhakaran, M.~Diaz, B.~Hutchinson,
  K.~Olson, A.~Molina, E.~Hoffman{-}John, J.~Lee, L.~Aroyo, R.~Rajakumar,
  A.~Butryna, M.~Lamm, V.~Kuzmina, J.~Fenton, A.~Cohen, R.~Bernstein,
  R.~Kurzweil, B.~Aguera{-}Arcas, C.~Cui, M.~Croak, E.~H. Chi, and Q.~Le,
  ``Lamda: Language models for dialog applications,'' \emph{CoRR}, vol.
  abs/2201.08239, 2022. [Online]. Available:
  \url{https://arxiv.org/abs/2201.08239}
\BIBentrySTDinterwordspacing

\bibitem{DBLP:journals/tcsv/ChengJWZ22}
\BIBentryALTinterwordspacing
X.~Cheng, M.~Jia, Q.~Wang, and J.~Zhang, ``A simple visual-textual baseline for
  pedestrian attribute recognition,'' \emph{{IEEE} Trans. Circuits Syst. Video
  Technol.}, vol.~32, no.~10, pp. 6994--7004, 2022. [Online]. Available:
  \url{https://doi.org/10.1109/TCSVT.2022.3178144}
\BIBentrySTDinterwordspacing

\bibitem{DBLP:journals/tcsv/HeP20}
\BIBentryALTinterwordspacing
X.~He and Y.~Peng, ``Fine-grained visual-textual representation learning,''
  \emph{{IEEE} Trans. Circuits Syst. Video Technol.}, vol.~30, no.~2, pp.
  520--531, 2020. [Online]. Available:
  \url{https://doi.org/10.1109/TCSVT.2019.2892802}
\BIBentrySTDinterwordspacing

\bibitem{DBLP:conf/nips/LuMX0CZTCK22}
\BIBentryALTinterwordspacing
P.~Lu, S.~Mishra, T.~Xia, L.~Qiu, K.~Chang, S.~Zhu, O.~Tafjord, P.~Clark, and
  A.~Kalyan, ``Learn to explain: Multimodal reasoning via thought chains for
  science question answering,'' in \emph{NeurIPS}, 2022. [Online]. Available:
  \url{http://papers.nips.cc/paper_files/paper/2022/hash/11332b6b6cf4485b84afadb1352d3a9a-Abstract-Conference.html}
\BIBentrySTDinterwordspacing

\bibitem{DBLP:journals/corr/abs-2302-00923}
\BIBentryALTinterwordspacing
Z.~Zhang, A.~Zhang, M.~Li, H.~Zhao, G.~Karypis, and A.~Smola, ``Multimodal
  chain-of-thought reasoning in language models,'' \emph{CoRR}, vol.
  abs/2302.00923, 2023. [Online]. Available:
  \url{https://doi.org/10.48550/arXiv.2302.00923}
\BIBentrySTDinterwordspacing

\bibitem{DBLP:journals/corr/abs-2301-13379}
\BIBentryALTinterwordspacing
Q.~Lyu, S.~Havaldar, A.~Stein, L.~Zhang, D.~Rao, E.~Wong, M.~Apidianaki, and
  C.~Callison{-}Burch, ``Faithful chain-of-thought reasoning,'' \emph{CoRR},
  vol. abs/2301.13379, 2023. [Online]. Available:
  \url{https://doi.org/10.48550/arXiv.2301.13379}
\BIBentrySTDinterwordspacing

\bibitem{lin2014microsoft}
T.-Y. Lin, M.~Maire, S.~Belongie, J.~Hays, P.~Perona, D.~Ramanan,
  P.~Doll{\'a}r, and C.~L. Zitnick, ``Microsoft coco: Common objects in
  context,'' in \emph{Computer Vision--ECCV 2014: 13th European Conference,
  Zurich, Switzerland, September 6-12, 2014, Proceedings, Part V 13}.\hskip 1em
  plus 0.5em minus 0.4em\relax Springer, 2014, pp. 740--755.

\bibitem{DBLP:journals/tcsv/XieFCHL22}
\BIBentryALTinterwordspacing
J.~Xie, W.~Fang, Y.~Cai, Q.~Huang, and Q.~Li, ``Knowledge-based visual question
  generation,'' \emph{{IEEE} Trans. Circuits Syst. Video Technol.}, vol.~32,
  no.~11, pp. 7547--7558, 2022. [Online]. Available:
  \url{https://doi.org/10.1109/TCSVT.2022.3189242}
\BIBentrySTDinterwordspacing

\bibitem{DBLP:journals/tcsv/WangWLLZW23}
\BIBentryALTinterwordspacing
Y.~Wang, B.~Wei, J.~Liu, Q.~Lin, L.~Zhang, and Y.~Wu, ``Spatial-semantic
  collaborative graph network for textbook question answering,'' \emph{{IEEE}
  Trans. Circuits Syst. Video Technol.}, vol.~33, no.~7, pp. 3214--3228, 2023.
  [Online]. Available: \url{https://doi.org/10.1109/TCSVT.2022.3231463}
\BIBentrySTDinterwordspacing

\bibitem{DBLP:journals/tcsv/DavaniS21}
\BIBentryALTinterwordspacing
S.~G. Davani and N.~J. Sarhan, ``Experimental analysis of optimal bandwidth
  allocation in computer vision systems,'' \emph{{IEEE} Trans. Circuits Syst.
  Video Technol.}, vol.~31, no.~10, pp. 4121--4130, 2021. [Online]. Available:
  \url{https://doi.org/10.1109/TCSVT.2020.3044015}
\BIBentrySTDinterwordspacing

\bibitem{gao2022simvp}
Z.~Gao, C.~Tan, L.~Wu, and S.~Z. Li, ``Simvp: Simpler yet better video
  prediction,'' in \emph{Proceedings of the IEEE/CVF Conference on Computer
  Vision and Pattern Recognition}, 2022, pp. 3170--3180.

\bibitem{tan2022simvp}
C.~Tan, Z.~Gao, S.~Li, and S.~Z. Li, ``Simvp: Towards simple yet powerful
  spatiotemporal predictive learning,'' \emph{arXiv preprint arXiv:2211.12509},
  2022.

\bibitem{cao2022survey}
H.~Cao, C.~Tan, Z.~Gao, G.~Chen, P.-A. Heng, and S.~Z. Li, ``A survey on
  generative diffusion model,'' \emph{arXiv preprint arXiv:2209.02646}, 2022.

\bibitem{DBLP:journals/tcsv/JinZFZLLZ22}
\BIBentryALTinterwordspacing
J.~Jin, X.~Zhang, X.~Fu, H.~Zhang, W.~Lin, J.~Lou, and Y.~Zhao, ``Just
  noticeable difference for deep machine vision,'' \emph{{IEEE} Trans. Circuits
  Syst. Video Technol.}, vol.~32, no.~6, pp. 3452--3461, 2022. [Online].
  Available: \url{https://doi.org/10.1109/TCSVT.2021.3113572}
\BIBentrySTDinterwordspacing

\bibitem{zheng2023cvt}
J.~Zheng, Y.~Wang, C.~Tan, S.~Li, G.~Wang, J.~Xia, Y.~Chen, and S.~Z. Li,
  ``Cvt-slr: Contrastive visual-textual transformation for sign language
  recognition with variational alignment,'' in \emph{Proceedings of the
  IEEE/CVF Conference on Computer Vision and Pattern Recognition}, 2023, pp.
  23\,141--23\,150.

\bibitem{DBLP:conf/iccv/AntolALMBZP15}
\BIBentryALTinterwordspacing
S.~Antol, A.~Agrawal, J.~Lu, M.~Mitchell, D.~Batra, C.~L. Zitnick, and
  D.~Parikh, ``{VQA:} visual question answering,'' in \emph{2015 {IEEE}
  International Conference on Computer Vision, {ICCV} 2015, Santiago, Chile,
  December 7-13, 2015}.\hskip 1em plus 0.5em minus 0.4em\relax {IEEE} Computer
  Society, 2015, pp. 2425--2433. [Online]. Available:
  \url{https://doi.org/10.1109/ICCV.2015.279}
\BIBentrySTDinterwordspacing

\bibitem{WU201721}
\BIBentryALTinterwordspacing
Q.~Wu, D.~Teney, P.~Wang, C.~Shen, A.~Dick, and A.~{van den Hengel}, ``Visual
  question answering: A survey of methods and datasets,'' \emph{Computer Vision
  and Image Understanding}, vol. 163, pp. 21--40, 2017, language in Vision.
  [Online]. Available:
  \url{https://www.sciencedirect.com/science/article/pii/S1077314217300772}
\BIBentrySTDinterwordspacing

\bibitem{DBLP:journals/tmm/YuZLQHTW20}
\BIBentryALTinterwordspacing
J.~Yu, W.~Zhang, Y.~Lu, Z.~Qin, Y.~Hu, J.~Tan, and Q.~Wu, ``Reasoning on the
  relation: Enhancing visual representation for visual question answering and
  cross-modal retrieval,'' \emph{{IEEE} Trans. Multim.}, vol.~22, no.~12, pp.
  3196--3209, 2020. [Online]. Available:
  \url{https://doi.org/10.1109/TMM.2020.2972830}
\BIBentrySTDinterwordspacing

\bibitem{DBLP:conf/icip/ChowdhuryNFS17}
\BIBentryALTinterwordspacing
I.~Chowdhury, K.~Nguyen, C.~Fookes, and S.~Sridharan, ``A cascaded long
  short-term memory {(LSTM)} driven generic visual question answering
  {(VQA)},'' in \emph{2017 {IEEE} International Conference on Image Processing,
  {ICIP} 2017, Beijing, China, September 17-20, 2017}.\hskip 1em plus 0.5em
  minus 0.4em\relax {IEEE}, 2017, pp. 1842--1846. [Online]. Available:
  \url{https://doi.org/10.1109/ICIP.2017.8296600}
\BIBentrySTDinterwordspacing

\bibitem{DBLP:conf/icip/HuangKJLJT18}
\BIBentryALTinterwordspacing
L.~Huang, K.~Kulkarni, A.~Jha, S.~Lohit, S.~Jayasuriya, and P.~K. Turaga,
  ``{CS-VQA:} visual question answering with compressively sensed images,'' in
  \emph{2018 {IEEE} International Conference on Image Processing, {ICIP} 2018,
  Athens, Greece, October 7-10, 2018}.\hskip 1em plus 0.5em minus 0.4em\relax
  {IEEE}, 2018, pp. 1283--1287. [Online]. Available:
  \url{https://doi.org/10.1109/ICIP.2018.8451445}
\BIBentrySTDinterwordspacing

\bibitem{DBLP:journals/tmm/Al-HalahG21}
\BIBentryALTinterwordspacing
Z.~Al{-}Halah and K.~Grauman, ``Modeling fashion influence from photos,''
  \emph{{IEEE} Trans. Multim.}, vol.~23, pp. 4143--4157, 2021. [Online].
  Available: \url{https://doi.org/10.1109/TMM.2020.3037459}
\BIBentrySTDinterwordspacing

\bibitem{DBLP:journals/tcsv/OuCW21}
\BIBentryALTinterwordspacing
Y.~Ou, Z.~Chen, and F.~Wu, ``Multimodal local-global attention network for
  affective video content analysis,'' \emph{{IEEE} Trans. Circuits Syst. Video
  Technol.}, vol.~31, no.~5, pp. 1901--1914, 2021. [Online]. Available:
  \url{https://doi.org/10.1109/TCSVT.2020.3014889}
\BIBentrySTDinterwordspacing

\bibitem{DBLP:conf/miccai/SeenivasanIKR22}
\BIBentryALTinterwordspacing
L.~Seenivasan, M.~Islam, A.~K. Krishna, and H.~Ren, ``Surgical-vqa: Visual
  question answering in surgical scenes using transformer,'' in \emph{Medical
  Image Computing and Computer Assisted Intervention - {MICCAI} 2022 - 25th
  International Conference, Singapore, September 18-22, 2022, Proceedings, Part
  {VII}}, ser. Lecture Notes in Computer Science, L.~Wang, Q.~Dou, P.~T.
  Fletcher, S.~Speidel, and S.~Li, Eds., vol. 13437.\hskip 1em plus 0.5em minus
  0.4em\relax Springer, 2022, pp. 33--43. [Online]. Available:
  \url{https://doi.org/10.1007/978-3-031-16449-1_4}
\BIBentrySTDinterwordspacing

\bibitem{DBLP:journals/tcsv/TangLJPZDK23}
\BIBentryALTinterwordspacing
J.~Tang, D.~Liu, X.~Jin, Y.~Peng, Q.~Zhao, Y.~Ding, and W.~Kong, ``{BAFN:}
  bi-direction attention based fusion network for multimodal sentiment
  analysis,'' \emph{{IEEE} Trans. Circuits Syst. Video Technol.}, vol.~33,
  no.~4, pp. 1966--1978, 2023. [Online]. Available:
  \url{https://doi.org/10.1109/TCSVT.2022.3218018}
\BIBentrySTDinterwordspacing

\bibitem{DBLP:journals/frai/WellsB21}
\BIBentryALTinterwordspacing
L.~Wells and T.~Bednarz, ``Explainable {AI} and reinforcement learning - {A}
  systematic review of current approaches and trends,'' \emph{Frontiers Artif.
  Intell.}, vol.~4, p. 550030, 2021. [Online]. Available:
  \url{https://doi.org/10.3389/frai.2021.550030}
\BIBentrySTDinterwordspacing

\bibitem{DBLP:journals/tcsv/WenP21}
\BIBentryALTinterwordspacing
Z.~Wen and Y.~Peng, ``Multi-level knowledge injecting for visual commonsense
  reasoning,'' \emph{{IEEE} Trans. Circuits Syst. Video Technol.}, vol.~31,
  no.~3, pp. 1042--1054, 2021. [Online]. Available:
  \url{https://doi.org/10.1109/TCSVT.2020.2991866}
\BIBentrySTDinterwordspacing

\bibitem{DBLP:journals/jair/RasXGD22}
\BIBentryALTinterwordspacing
G.~Ras, N.~Xie, M.~van Gerven, and D.~Doran, ``Explainable deep learning: {A}
  field guide for the uninitiated,'' \emph{J. Artif. Intell. Res.}, vol.~73,
  pp. 329--396, 2022. [Online]. Available:
  \url{https://doi.org/10.1613/jair.1.13200}
\BIBentrySTDinterwordspacing

\bibitem{DBLP:journals/tmm/WuLZLZQC23}
\BIBentryALTinterwordspacing
Y.~Wu, L.~Liao, G.~Zhang, W.~Lei, G.~Zhao, X.~Qian, and T.~Chua, ``State graph
  reasoning for multimodal conversational recommendation,'' \emph{{IEEE} Trans.
  Multim.}, vol.~25, pp. 3113--3124, 2023. [Online]. Available:
  \url{https://doi.org/10.1109/TMM.2022.3155900}
\BIBentrySTDinterwordspacing

\bibitem{DBLP:journals/tcsv/LiMDFT23}
\BIBentryALTinterwordspacing
W.~Li, Z.~Ma, L.~Deng, X.~Fan, and Y.~Tian, ``Neuron-based spiking transmission
  and reasoning network for robust image-text retrieval,'' \emph{{IEEE} Trans.
  Circuits Syst. Video Technol.}, vol.~33, no.~7, pp. 3516--3528, 2023.
  [Online]. Available: \url{https://doi.org/10.1109/TCSVT.2022.3233042}
\BIBentrySTDinterwordspacing

\bibitem{gao2023llamaadapterv2}
P.~Gao, J.~Han, R.~Zhang, Z.~Lin, S.~Geng, A.~Zhou, W.~Zhang, P.~Lu, C.~He,
  X.~Yue, H.~Li, and Y.~Qiao, ``Llama-adapter v2: Parameter-efficient visual
  instruction model,'' \emph{arXiv preprint arXiv:2304.15010}, 2023.

\bibitem{touvron2023llama}
H.~Touvron, T.~Lavril, G.~Izacard, X.~Martinet, M.-A. Lachaux, T.~Lacroix,
  B.~Rozi{\`e}re, N.~Goyal, E.~Hambro, F.~Azhar \emph{et~al.}, ``Llama: Open
  and efficient foundation language models,'' \emph{arXiv preprint
  arXiv:2302.13971}, 2023.

\bibitem{chen2015microsoft}
X.~Chen, H.~Fang, T.-Y. Lin, R.~Vedantam, S.~Gupta, P.~Doll{\'a}r, and C.~L.
  Zitnick, ``Microsoft coco captions: Data collection and evaluation server,''
  \emph{arXiv preprint arXiv:1504.00325}, 2015.

\bibitem{goyal2017making}
Y.~Goyal, T.~Khot, D.~Summers-Stay, D.~Batra, and D.~Parikh, ``Making the v in
  vqa matter: Elevating the role of image understanding in visual question
  answering,'' in \emph{Proceedings of the IEEE conference on computer vision
  and pattern recognition}, 2017, pp. 6904--6913.

\bibitem{zhu2023minigpt}
D.~Zhu, J.~Chen, X.~Shen, X.~Li, and M.~Elhoseiny, ``Minigpt-4: Enhancing
  vision-language understanding with advanced large language models,''
  \emph{arXiv preprint arXiv:2304.10592}, 2023.

\bibitem{DBLP:conf/nips/VaswaniSPUJGKP17}
\BIBentryALTinterwordspacing
A.~Vaswani, N.~Shazeer, N.~Parmar, J.~Uszkoreit, L.~Jones, A.~N. Gomez,
  L.~Kaiser, and I.~Polosukhin, ``Attention is all you need,'' in
  \emph{Advances in Neural Information Processing Systems 30: Annual Conference
  on Neural Information Processing Systems 2017, December 4-9, 2017, Long
  Beach, CA, {USA}}, I.~Guyon, U.~von Luxburg, S.~Bengio, H.~M. Wallach,
  R.~Fergus, S.~V.~N. Vishwanathan, and R.~Garnett, Eds., 2017, pp. 5998--6008.
  [Online]. Available:
  \url{https://proceedings.neurips.cc/paper/2017/hash/3f5ee243547dee91fbd053c1c4a845aa-Abstract.html}
\BIBentrySTDinterwordspacing

\bibitem{DBLP:conf/cvpr/Yu0CT019}
\BIBentryALTinterwordspacing
Z.~Yu, J.~Yu, Y.~Cui, D.~Tao, and Q.~Tian, ``Deep modular co-attention networks
  for visual question answering,'' in \emph{{IEEE} Conference on Computer
  Vision and Pattern Recognition, {CVPR} 2019, Long Beach, CA, USA, June 16-20,
  2019}.\hskip 1em plus 0.5em minus 0.4em\relax Computer Vision Foundation /
  {IEEE}, 2019, pp. 6281--6290. [Online]. Available:
  \url{http://openaccess.thecvf.com/content_CVPR_2019/html/Yu_Deep_Modular_Co-Attention_Networks_for_Visual_Question_Answering_CVPR_2019_paper.html}
\BIBentrySTDinterwordspacing

\bibitem{DBLP:conf/cvpr/00010BT0GZ18}
\BIBentryALTinterwordspacing
P.~Anderson, X.~He, C.~Buehler, D.~Teney, M.~Johnson, S.~Gould, and L.~Zhang,
  ``Bottom-up and top-down attention for image captioning and visual question
  answering,'' in \emph{2018 {IEEE} Conference on Computer Vision and Pattern
  Recognition, {CVPR} 2018, Salt Lake City, UT, USA, June 18-22, 2018}.\hskip
  1em plus 0.5em minus 0.4em\relax Computer Vision Foundation / {IEEE} Computer
  Society, 2018, pp. 6077--6086. [Online]. Available:
  \url{http://openaccess.thecvf.com/content_cvpr_2018/html/Anderson_Bottom-Up_and_Top-Down_CVPR_2018_paper.html}
\BIBentrySTDinterwordspacing

\bibitem{DBLP:journals/titb/YuPFCS23}
\BIBentryALTinterwordspacing
R.~Yu, C.~Pan, X.~Fei, M.~Chen, and D.~Shen, ``Multi-graph attention networks
  with bilinear convolution for diagnosis of schizophrenia,'' \emph{{IEEE} J.
  Biomed. Health Informatics}, vol.~27, no.~3, pp. 1443--1454, 2023. [Online].
  Available: \url{https://doi.org/10.1109/JBHI.2022.3229465}
\BIBentrySTDinterwordspacing

\bibitem{DBLP:conf/cvpr/GaoJYLHWL19}
\BIBentryALTinterwordspacing
P.~Gao, Z.~Jiang, H.~You, P.~Lu, S.~C.~H. Hoi, X.~Wang, and H.~Li, ``Dynamic
  fusion with intra- and inter-modality attention flow for visual question
  answering,'' in \emph{{IEEE} Conference on Computer Vision and Pattern
  Recognition, {CVPR} 2019, Long Beach, CA, USA, June 16-20, 2019}.\hskip 1em
  plus 0.5em minus 0.4em\relax Computer Vision Foundation / {IEEE}, 2019, pp.
  6639--6648. [Online]. Available:
  \url{http://openaccess.thecvf.com/content_CVPR_2019/html/Gao_Dynamic_Fusion_With_Intra-_and_Inter-Modality_Attention_Flow_for_Visual_CVPR_2019_paper.html}
\BIBentrySTDinterwordspacing

\bibitem{DBLP:conf/icml/KimSK21}
\BIBentryALTinterwordspacing
W.~Kim, B.~Son, and I.~Kim, ``Vilt: Vision-and-language transformer without
  convolution or region supervision,'' in \emph{Proceedings of the 38th
  International Conference on Machine Learning, {ICML} 2021, 18-24 July 2021,
  Virtual Event}, ser. Proceedings of Machine Learning Research, M.~Meila and
  T.~Zhang, Eds., vol. 139.\hskip 1em plus 0.5em minus 0.4em\relax {PMLR},
  2021, pp. 5583--5594. [Online]. Available:
  \url{http://proceedings.mlr.press/v139/kim21k.html}
\BIBentrySTDinterwordspacing

\bibitem{lu2021iconqa}
P.~Lu, L.~Qiu, J.~Chen, T.~Xia, Y.~Zhao, W.~Zhang, Z.~Yu, X.~Liang, and S.-C.
  Zhu, ``Iconqa: A new benchmark for abstract diagram understanding and visual
  language reasoning,'' in \emph{The 35th Conference on Neural Information
  Processing Systems (NeurIPS 2021) Track on Datasets and Benchmarks}, 2021.

\bibitem{li2019visualbert}
L.~H. Li, M.~Yatskar, D.~Yin, C.-J. Hsieh, and K.-W. Chang, ``Visualbert: A
  simple and performant baseline for vision and language,'' \emph{arXiv
  preprint arXiv:1908.03557}, 2019.

\bibitem{khashabi-etal-2020-unifiedqa}
\BIBentryALTinterwordspacing
D.~Khashabi, S.~Min, T.~Khot, A.~Sabharwal, O.~Tafjord, P.~Clark, and
  H.~Hajishirzi, ``{UNIFIEDQA}: Crossing format boundaries with a single {QA}
  system,'' in \emph{Findings of the Association for Computational Linguistics:
  EMNLP 2020}.\hskip 1em plus 0.5em minus 0.4em\relax Online: Association for
  Computational Linguistics, Nov. 2020, pp. 1896--1907. [Online]. Available:
  \url{https://aclanthology.org/2020.findings-emnlp.171}
\BIBentrySTDinterwordspacing

\bibitem{DBLP:journals/corr/abs-2304-08485}
\BIBentryALTinterwordspacing
H.~Liu, C.~Li, Q.~Wu, and Y.~J. Lee, ``Visual instruction tuning,''
  \emph{CoRR}, vol. abs/2304.08485, 2023. [Online]. Available:
  \url{https://doi.org/10.48550/arXiv.2304.08485}
\BIBentrySTDinterwordspacing

\bibitem{DBLP:journals/corr/abs-2210-11416}
\BIBentryALTinterwordspacing
H.~W. Chung, L.~Hou, S.~Longpre, B.~Zoph, Y.~Tay, W.~Fedus, E.~Li, X.~Wang,
  M.~Dehghani, S.~Brahma, A.~Webson, S.~S. Gu, Z.~Dai, M.~Suzgun, X.~Chen,
  A.~Chowdhery, S.~Narang, G.~Mishra, A.~Yu, V.~Y. Zhao, Y.~Huang, A.~M. Dai,
  H.~Yu, S.~Petrov, E.~H. Chi, J.~Dean, J.~Devlin, A.~Roberts, D.~Zhou, Q.~V.
  Le, and J.~Wei, ``Scaling instruction-finetuned language models,''
  \emph{CoRR}, vol. abs/2210.11416, 2022. [Online]. Available:
  \url{https://doi.org/10.48550/arXiv.2210.11416}
\BIBentrySTDinterwordspacing

\end{thebibliography}

% \newpage
% 
\section*{Biography Section}
\begin{IEEEbiography}[{\includegraphics[width=1in,height=1.25in,clip,keepaspectratio]{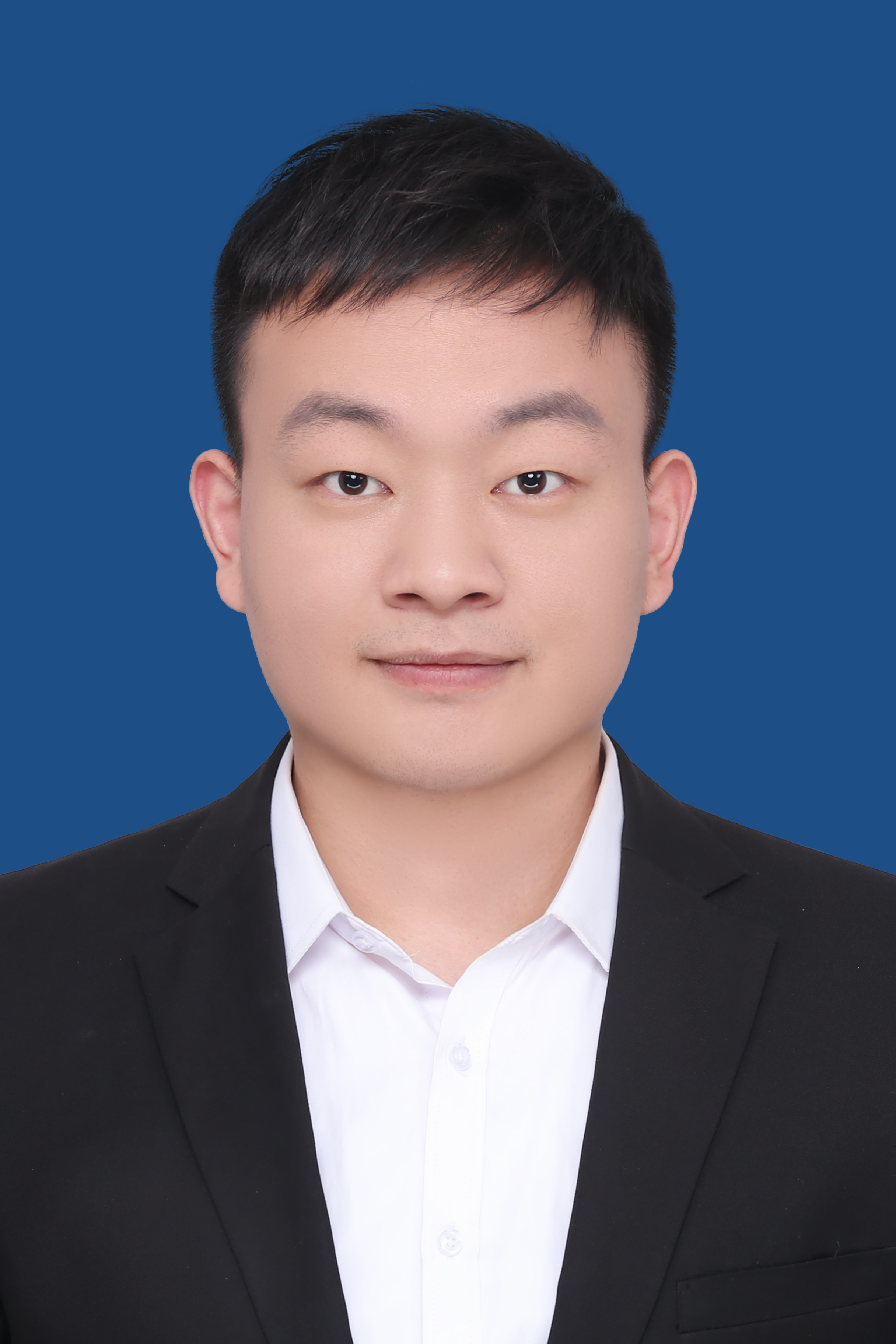}}]{Jingxuan Wei} is currently pursuing the Ph.D. degree at the University of Chinese Academy of Sciences, trained by Shenyang Institute of Computing Technology, Chinese Academy of Sciences. He has participated in multimodal research internships and academic exchanges at various companies and institutions, including Kuaishou Technology, ByteDance, and Microsoft Research Asia. Additionally, he has been actively involved in the practical application of numerous foundation models. His primary research interest lies in multimodal learning.
\end{IEEEbiography}
\begin{IEEEbiography}[{\includegraphics[width=1in,height=1.25in,clip,keepaspectratio]{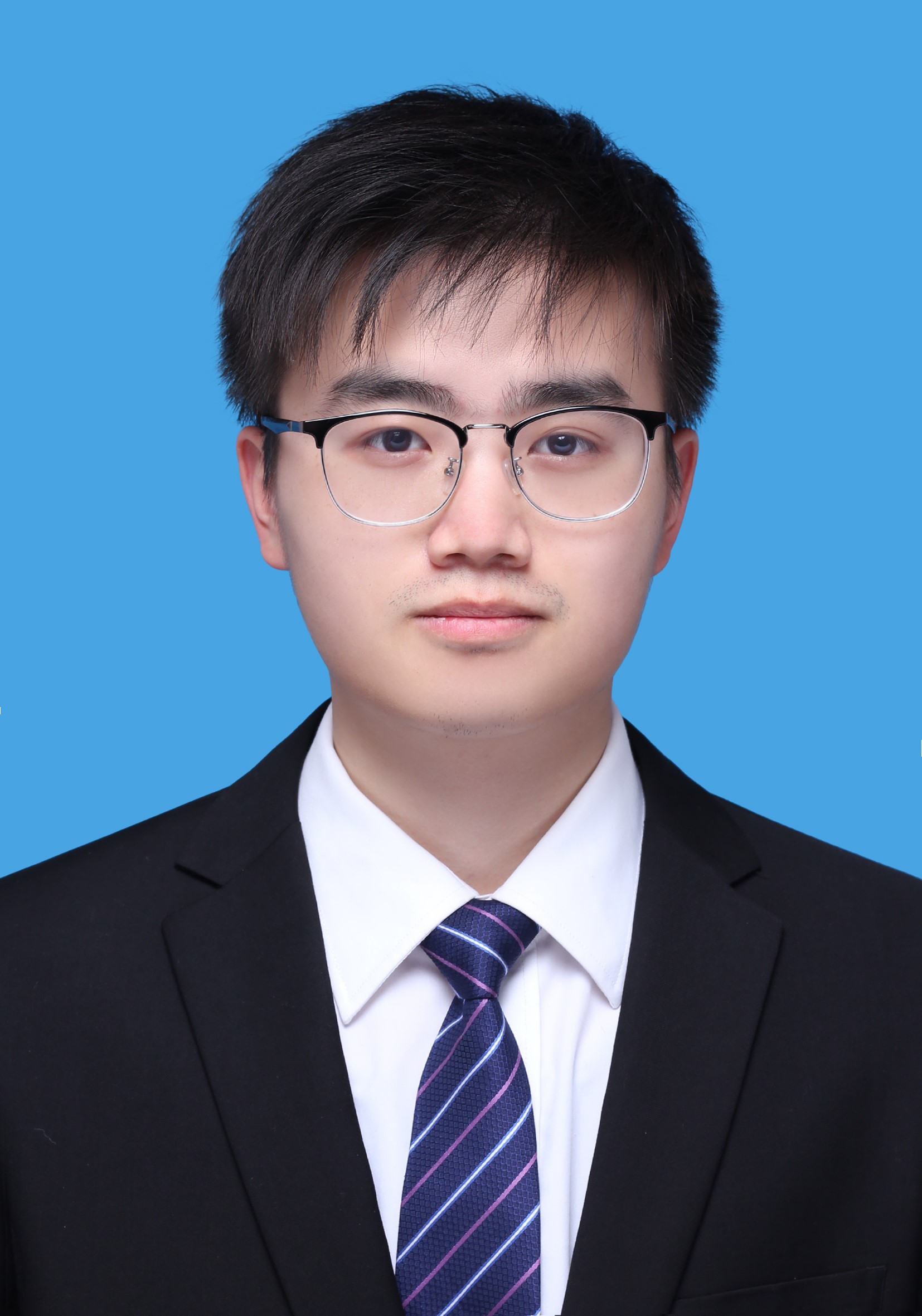}}]{Cheng Tan} is currently pursuing the Ph.D. degree with the School of Engineering, Westlake University, Hangzhou, China. He has authored several papers in top-tier conferences and journals, such as the IEEE Transactions on Knowledge and Data Engineering (TKDE), IEEE Transactions on Neural Networks and Learning Systems (TNNLS), ACM Multimedia (MM), International Conference on Learning Representations (ICLR), and Conference on Computer Vision and Pattern Recognition (CVPR). His main research interest is self-supervised learning.
\end{IEEEbiography}
\begin{IEEEbiography}[{\includegraphics[width=1in,height=1.25in,clip,keepaspectratio]{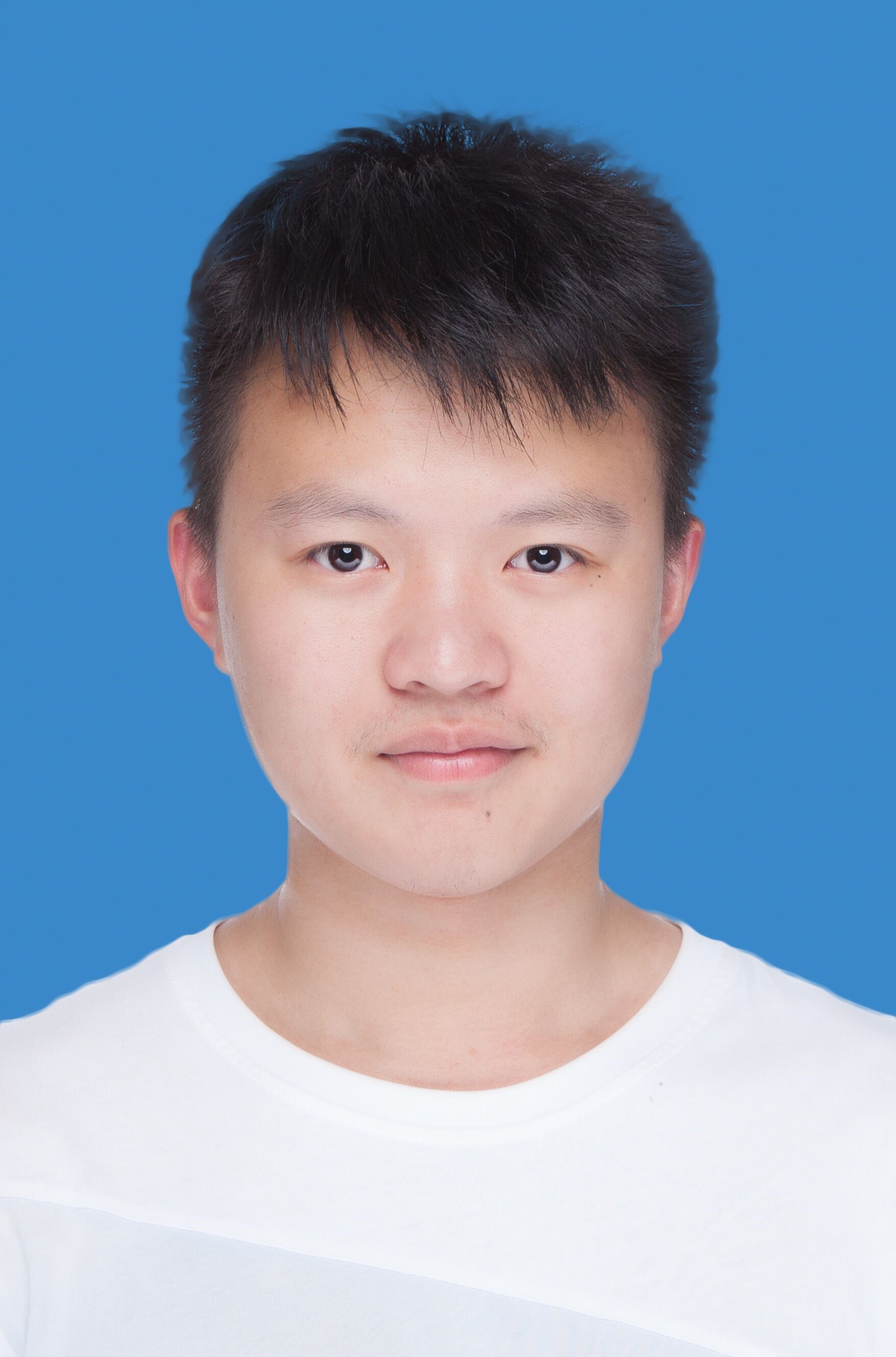}}]{Zhangyang Gao} is currently a Ph.D. candidate at the School of Engineering, Westlake University, Hangzhou, China. He has authored several papers in top-tier conferences and journals, such as the IEEE Transactions on Knowledge and Data Engineering (TKDE), IEEE Transactions on Neural Networks and Learning Systems (TNNLS), International Conference on Learning Representations (ICLR), and Conference on Computer Vision and Pattern Recognition (CVPR). His main research interests include unsupervised video tasks and drug discovery.
\end{IEEEbiography}
\begin{IEEEbiography}[{\includegraphics[width=1in,height=1.25in,clip,keepaspectratio]{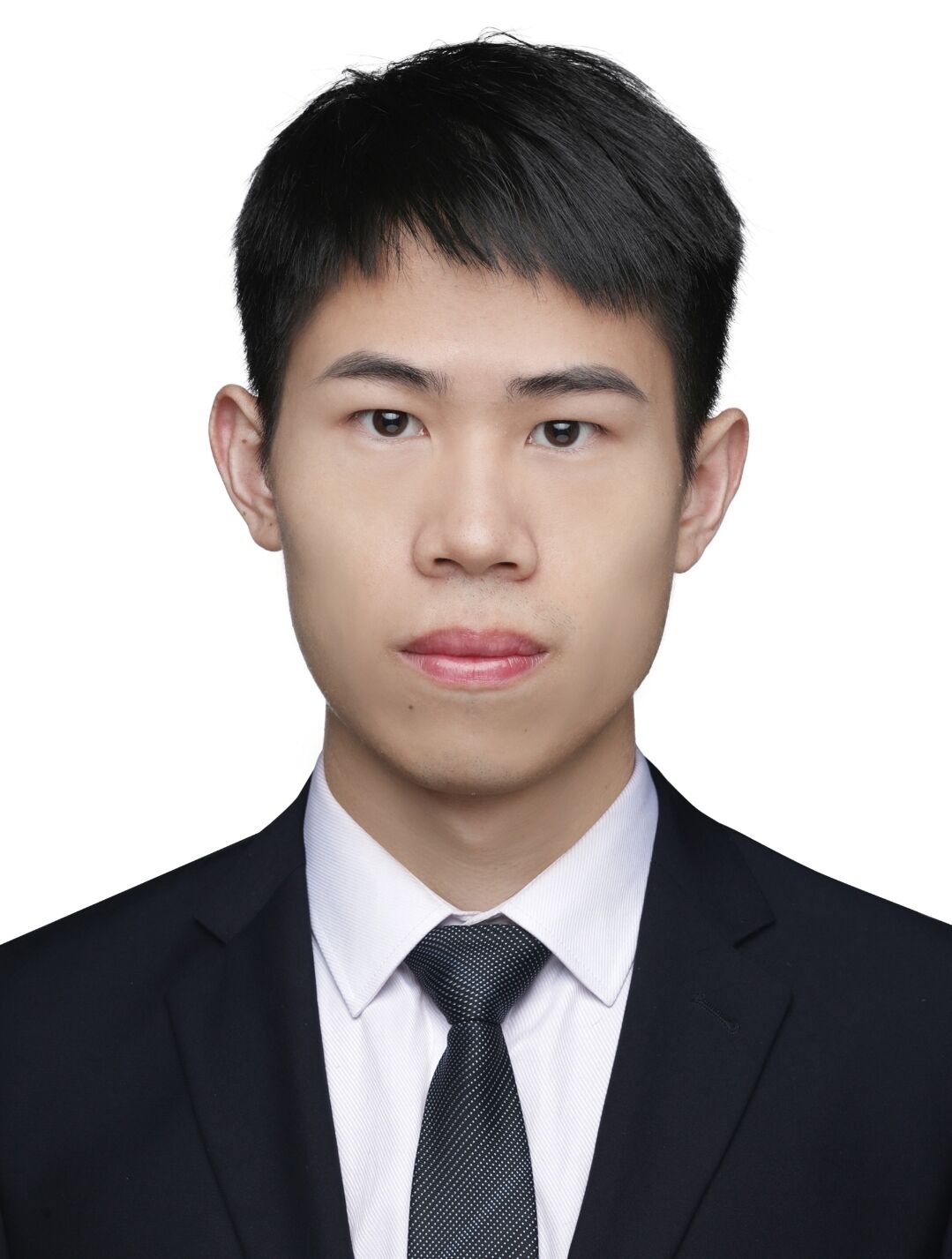}}]{Linzhuang Sun} is currently a Ph.D candidate at the University of Chinese Academy of Sciences, Beijing, China. He has authored several papers in high-level conference, like the International Conference on Advanced Data Mining and Applications(ADMA). He also has worked as an intern at BAAI and Baidu before and after, participating in large model development. And his main research interests include multimodal learning and sentiment analysis.
\end{IEEEbiography}
\begin{IEEEbiography}[{\includegraphics[width=1in,height=1.25in,clip,keepaspectratio]{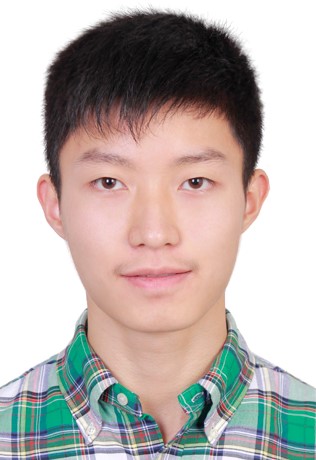}}]{Siyuan Li} received the B.S. degree from Nanjing University, Nanjing, China in 2021. He is currently pursuing the Ph.D. degree with the School of Engineering, Westlake University, Hangzhou. 
He has authored several papers in top-tier conferences, such as European Conference on Computer Vision (ECCV), and International Conference on Machine Learning (ICML). His main research interest is self-supervised learning.
\end{IEEEbiography}
% \vspace{1pt}
\begin{IEEEbiography}[{\includegraphics[width=1in,height=1.25in,clip,keepaspectratio]{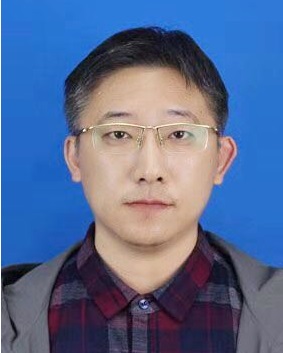}}]{Bihui Yu} is a researcher and doctoral advisor at the University of Chinese Academy of Sciences, Beijing, China. He has presided over and participated in many provincial and national important projects, like the Network culture market dynamic supervision service system research and development and application demonstration. He also has authored several high level papers and invention patents. And his research interests include knowledge engineering, big data and multimodal learning.
\end{IEEEbiography}
% \vspace{1pt}
\begin{IEEEbiography}[{\includegraphics[width=1in,height=1.25in,clip,keepaspectratio]{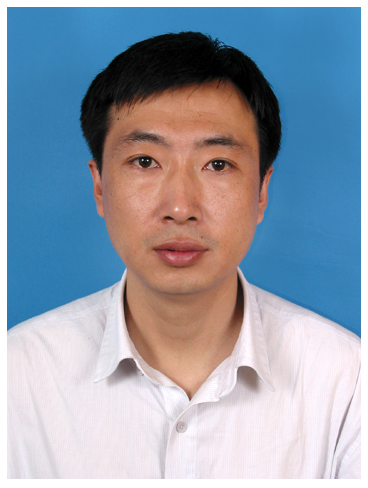}}]{Ruifeng Guo} , PhD supervisor, is the director of Shenyang Institute of Computing Technology, Chinese Academy of Sciences. He has served as the head of numerous projects, including the development of advanced numerical control devices based on the domestic "Loongson" CPU chip and the establishment of the National Open Numerical Control System Supporting Technology Innovation Platform. His exceptional contributions cater to the major strategic and industrial needs of the nation. In the past two years, he has led his team to achieve numerous accolades, including the prestigious National Science and Technology Progress Second Prize. His research interests encompass machine learning, high-end numerical control, and the application of various cross-disciplinary fields related to artificial intelligence.
\end{IEEEbiography}
% \vspace{1pt}
\begin{IEEEbiography}[{\includegraphics[width=1in,height=1.25in,clip,keepaspectratio]{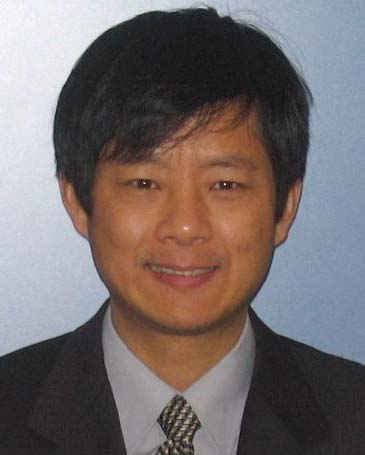}}]{Stan Z. Li} (Fellow, IEEE) received the B.Eng. degree from Hunan University, China, the M.Eng. degree from the National University of Defense Technology, China, and the Ph.D. degree from the University of Surrey, U.K. He was an Associate Professor (tenure) at Nanyang Technological University, Singapore. He was with Microsoft Research Asia, Beijing, China, as a Research Lead, from 2000 to 2004. He was the Director of the Center for Biometrics and Security Research (CBSR), Chinese Academy of Sciences, Beijing, from 2004 to 2019. He joined Westlake University as a Chair Professor of artificial intelligence in February 2019. He has authored over 400 papers in international journals and conferences and has authored and edited ten books, with over 60,000 Google Scholar citations. Among these are Markov Random Field Models in Image Analysis (Springer), Handbook of Face Recognition (Springer), and Encyclopedia of Biometrics (Springer). His research interests include fundamental research in machine learning, data science, and applied research in multiple AI-related interdisciplinary fields (computer vision, smart sensors, life science, material science, and environmental science).

Dr. Li served as an Associate Editor of the IEEE Transactions on Pattern Analysis and Machine Intelligence (TPAMI) and organized more than 100 international conferences or workshops.
\end{IEEEbiography}

 

% \end{IEEEbiography}



% \bibliographystyle{acl_natbib}

% \vfill

\end{document}